\documentclass{article}
\usepackage[preprint]{neurips_2026}

\usepackage[utf8]{inputenc} % allow utf-8 input
\usepackage[T1]{fontenc}    % use 8-bit T1 fonts
\usepackage{hyperref}       % hyperlinks
\usepackage{url}            % simple URL typesetting
\usepackage{booktabs}       % professional-quality tables
\usepackage{amsfonts}       % blackboard math symbols
\usepackage{nicefrac}       % compact symbols for 1/2, etc.
\usepackage{microtype}      % microtypography
\usepackage{xcolor}         % colors
\usepackage{array}    % For better column formatting
\usepackage{caption}
\usepackage{tabularx}
\usepackage{rotating}
\usepackage{tcolorbox}
\usepackage{algorithm}
\usepackage{algpseudocode}
\usepackage{amsmath}
\usepackage{algorithmicx}
\usepackage{booktabs}
\usepackage{multirow}
\usepackage{multicol}
\usepackage{amsfonts}
\usepackage{cleveref}
\usepackage{tabularx}
\usepackage{bbm}
\usepackage{wrapfig}
\usepackage{amssymb}
\usepackage{enumitem}
\setlist[itemize]{itemsep=0pt, topsep=0pt, parsep=0pt, partopsep=0pt}
\usepackage{graphicx}
\usepackage{subcaption}
\title{Test-Time Compositional Generalization \\in Diffusion Models via Concept Discovery}

\author{
    Zekun Wang$^{1}$ $\quad$ Anant Gupta$^{1}$ $\quad$ Tianyi Zhu$^{2}$ $\quad$ Christopher J. MacLellan$^{1}$ \\
    $^{1}$Georgia Institute of Technology \quad $^{2}$University of Virginia \\
    \texttt{\{zekun, agupta886, cmaclell\}@gatech.edu} $\quad$
    \texttt{crv5ns@virginia.edu}
}

\begin{document}

\maketitle

\begin{abstract}
\label{sec:abs}
Compositional generalization requires models to produce novel configurations from familiar parts.
In diffusion models, prior compositional generation methods typically assume that the relevant concepts or conditioning signals are already available. 
We instead ask whether a pretrained diffusion model can discover query-specific concepts from the time-indexed scores it learns for the noisy marginals $p_t(x_t)$ and compose them at test time.
Given a single out-of-distribution query, our method performs gradient ascent on $s_\theta(x_t,t) \approx \nabla_{x_t}\log p_t(x_t)$ at multiple noising timesteps to recover local density modes, maps these modes into clean-space Gaussians, greedily selects relevant prototypes with a submodular likelihood objective, and combines them into a product-of-experts (PoE) teacher model with an analytic score.
This teacher model can be sampled directly through classifier-free guidance or used to generate a sample pool for training a new class embedding and low-rank adapter.
On held-out composition benchmarks built from ColorMNIST and CelebA, both the analytic PoE sampler and the low-rank adapted model outperform query-only and nearest trained-class baselines. 
These results suggest that the time-indexed score geometry of the diffusion model contains reusable density-mode concepts that support test-time compositional generation without a predefined concept library.
\end{abstract}

% \begin{abstract}
% Compositional generalization requires models to produce novel configurations from familiar parts. 
% In diffusion models, prior compositional generation methods typically assume that the relevant concepts or conditioning signals are already available. 
% We instead ask whether a pretrained diffusion model can discover query-specific primitives from its learned density and compose them at test time. 
% Given a single out-of-distribution query, our method performs gradient ascent at multiple noising timesteps on the marginal distribution $p_t(x_t)$ to recover local modes, maps these modes into clean-space Gaussian prototypes, greedily selects the most relevant prototypes by a submodular likelihood objective, and composes them into a product-of-experts teacher with an analytic score. 
% This analytic teacher can be sampled directly through classifier-free guidance, or used to generate a pool of samples for training a new class embedding and LoRA adapter. 
% On held-out composition benchmarks built from ColorMNIST and CelebA, both the analytic PoE sampler and the LoRA-adapted model outperform query-only and nearest trained-class baselines. 
% These results suggest that DDPM score geometry contains reusable prototypes that support test-time compositional generation without a predefined concept library.
% \end{abstract}
\section{Introduction}
\label{sec:intro}

Human intelligence is strikingly compositional.
After learning familiar parts, relations, or words, people can recombine them to understand and produce novel configurations. 
This ability has long motivated researchers in cognitive science and linguistics.
In these fields, systematic generalization is often linked to reusable primitives and operators for recombination \cite{fodor1988connectionism,lake2017building}.
It is also a central challenge for deep learning, where neural networks may interpolate well within a training distribution but fail when familiar primitives appear in unfamiliar combinations \cite{lake2018generalization,ruis2020benchmark}. 
The key question is therefore whether models can not only learn useful representations, but also discover reusable units and compose them at test time.
Deep learning researchers often address this problem by designing representations or architectures that support composition.
For example, neural module networks assemble learned components from linguistic or program structure \cite{andreas2016neural,andreas2016learning}; benchmarks such as SCAN test systematic recombination \cite{lake2018generalization,ruis2020benchmark}; and visual generation methods analyze whether learned objects, attributes, or concepts can be recombined out of distribution. 
While these directions have made important progress, they often assume that the relevant primitives are supplied by the benchmark, labels, prompts, or a predefined concept library.

A probabilistic view offers a complementary account.
Concepts can be treated as high-probability regions in an energy landscape, and composition as satisfying multiple constraints at once. 
This idea appears in Hopfield attractors and mixed states \cite{hopfield1982neural,amit1985spin}, product-of-experts models (PoE) \cite{hinton2002training}, and recent energy-based or diffusion composition methods \cite{du2020compositional,liu2022compositional,du2023reduce}. 
However, these methods typically assume that the component distributions or conditioning signals are given.
This leaves a gap between concept discovery and compositional generalization.
In open-ended visual domains, the relevant primitive may depend on the query and emerge at different levels of abstraction, such as digit identity, color, texture, or facial attributes.
A fuller account should therefore discover reusable concepts from the model's learned density function and compose them without fixed vocabularies or external labels.

Diffusion models provide a natural setting for this view.
A pretrained diffusion model learns time-indexed scores $s_\theta(x_t,t) \approx \nabla_{x_t}\log p_t(x_t)$ for the noisy marginals $p_t(x_t)$.
Rather than using these scores only for reverse sampling, we repurpose them as a hierarchy of density modes: different noise levels smooth the data distribution at different scales, so local modes of $p_t$ can act as discrete centroids at different levels of abstraction.
% Starting from a single out-of-distribution query, we hypothesize that ascending the learned scores of these noisy marginals across multiple timesteps recovers a neighborhood of local density modes whose clean-space projections encode reusable concept prototypes for the held-out composition.
% Rather than using these scores only for reverse sampling, we repurpose the noisy marginals as an implicit hierarchy of density modes: by ascending the learned scores across multiple timesteps from a single out-of-distribution query, we recover local modes whose clean-space projections act as reusable concept prototypes for the held-out composition.
This view is motivated by the density-mode hierarchy illustrated in Appendix~\ref{app:fmnist}, where modes of diffusion marginals form discrete clusters that coarsen across noise levels.
Starting from a single out-of-distribution query, our method maps each discovered mode into a clean-space local Gaussian, greedily selects the modes most relevant to the query with a submodular likelihood objective, and combines them into a PoE teacher model whose closed-form score can guide classifier-free diffusion sampling.
We also study an optional distillation step that uses samples from the PoE to train a new class embedding and low-rank adapter, thereby absorbing the discovered composition back into the base model's score manifold.
On ColorMNIST and CelebA, the discovered concept prototypes and their PoE composition outperform query-only sampling and nearest trained-class retrieval, with LoRA distillation further improving in-manifold generation.
% \paragraph{Contributions.}
% \begin{itemize}[leftmargin=*]
%     \item We bridge concept discovery and compositional generalization by repurposing a pretrained DDPM as an implicit hierarchy of density modes. Concept discovery becomes score-based mode finding, allowing prototypes to be recovered at test time without a predefined primitive pool.
%     \item We introduce a test-time composition method that greedily selects relevant discovered modes using a submodular likelihood objective and combines their local Gaussian approximations into a product-of-experts teacher with a closed-form score.
%     \item On two compositional benchmarks, ColorMNIST and CelebA, we show that mode-discovered prototypes and PoE composition encode richer information about unseen compositions than query-only or nearest trained-class baselines. The optional LoRA distillation step further improves in-manifold generation by learning the discovered composition back into the diffusion model.
% \end{itemize}
\paragraph{Contributions.}
\begin{itemize}[leftmargin=*]
    \item We bridge concept discovery and compositional generalization by repurposing a pretrained diffusion model as an implicit hierarchy of time-indexed density modes. Concept discovery becomes score-based mode finding over the noisy marginals $p_t(x_t)$, allowing discrete concept prototypes to be recovered at test time without a predefined primitive pool.
    \item We introduce a test-time composition method that greedily selects relevant discovered modes using a submodular likelihood objective and combines their clean-space local Gaussian approximations into a PoE teacher model with a closed-form score.
    \item On two compositional benchmarks, ColorMNIST and CelebA, we show that discovered concept prototypes and PoE composition encode richer information about unseen compositions than query-only or nearest trained-class baselines. The optional low-rank distillation step further improves in-manifold generation by learning the discovered composition back into the diffusion model.
\end{itemize}

\section{Related work}
\label{sec:relatedwork}

\paragraph{Concept discovery}

Concept discovery aims to identify reusable abstractions between pixels and labels, such as parts, attributes, or object-level factors. Prior work has studied such concepts as interpretable directions in representation space \citep{bau2017network,kim2018tcav,ghorbani2019ace}, as discriminative prototypes \citep{chen2019looks}, or as composable generative units \citep{du2020compositional,Liu2023}. These methods reveal meaningful visual structure, but are often discriminative, rely on external supervision, or do not directly expose the hierarchy of concepts.

From a statistical view, concepts can be interpreted as local density modes, with examples assigned to their basins of attraction. This connects concept discovery to mode-seeking and hierarchical clustering methods such as mean-shift \citep{comaniciu2002mean}, density-derivative-ratio mode estimation \citep{sasaki2018mode}, cluster trees \citep{chaudhuri2010rates}, and hierarchical prototype models \citep{wang2025neuripsdeep}. We instead recover concepts from the hierarchy already implicitly encoded by a pretrained diffusion model. Diffusion models admit a deep latent-variable interpretation across noise levels \citep{sohl2015deep,ho2020denoising,kingma2021variational,sonderby2016ladder,vahdat2020nvae}, and recent work shows that diffusion timesteps organize information from coarse semantic structure to fine detail, with phase-transition behavior at intermediate noise levels \citep{sclocchi2025phase, huang2024dreamtime}. This motivates our method: we discover concept prototypes by seeking modes of the smoothed marginal $p_t(x)$ using the pretrained score $\nabla_x \log p_t(x)$, requiring no additional training, external supervision, or manually specified taxonomy.

\paragraph{Compositional generalization}

Compositional generalization is the ability to recombine familiar primitives into novel configurations, a central problem in cognitive science, linguistics, and machine learning \citep{fodor1988connectionism,lake2017building,lake2018generalization}. In AI, this ability has been studied through diagnostic benchmarks such as SCAN and gSCAN \citep{lake2018generalization,ruis2020benchmark}, modular architectures that compose learned neural components \citep{andreas2016neural,andreas2016learning}, and data-augmentation or semantic-parsing methods that encourage systematic recombination \citep{andreas2020good,yin2021compositional}. These approaches differ in implementation, but typically share a common assumption: the relevant primitives are already specified by the benchmark, the language, or a predefined module set.

A complementary line of work provides a probabilistic framework for composition. Product-of-experts (PoE) models combine constraints by multiplying concept densities \citep{hinton2002training}; energy-based models use this principle to compose visual concepts \citep{du2020compositional}; and composable diffusion models implement analogous composition by adding concept-specific scores \citep{liu2022compositional}, with later work improving sampling through MCMC corrections \citep{du2023reduce}. Classifier-free guidance provides a related score-combination mechanism for conditional generation \citep{ho2022classifier}. These methods give a  mathematical account of how concepts can be composed, but still assume that the component distributions or conditioning signals are available beforehand.

Our work connects concept discovery with compositional generation, following the view that systematic generalization requires both reusable parts and an operator for recombining them \citep{fodor1988connectionism,lake2017building}. Closest to our setting, \cite{Liu2023} learns a shared library of concept embeddings from unlabeled images, whereas we discover query-specific primitives at test time as modes of the unconditional diffusion density and compose their local Gaussian approximations without a predefined concept vocabulary.
\paragraph{Diffusion models}

Diffusion models learn time-indexed score fields for the noisy marginals $p_t(x)$ and are typically used as generative samplers \citep{vincent2011connection,song2019generative,ho2020denoising,song2021score}. We repurpose this learned score geometry for concept discovery. Instead of immediately running the reverse sampler, we use $s_\theta(x,t) \approx \nabla_x \log p_t(x)$ to seek modes of intermediate marginals. This view is supported by recent analyses showing that diffusion timesteps encode structure across scales, including phase-transition behavior in which high-level class identity disappears sharply while lower-level features persist \citep{sclocchi2025phase}, suggesting that modes of $p_t$ can serve as concept prototypes at the granularity determined by~$t$.
\section{Discovering Concepts for Product-of-Experts Composition}
\label{sec:method}

\begin{figure}
    \centering
    \includegraphics[width=1.01\linewidth]{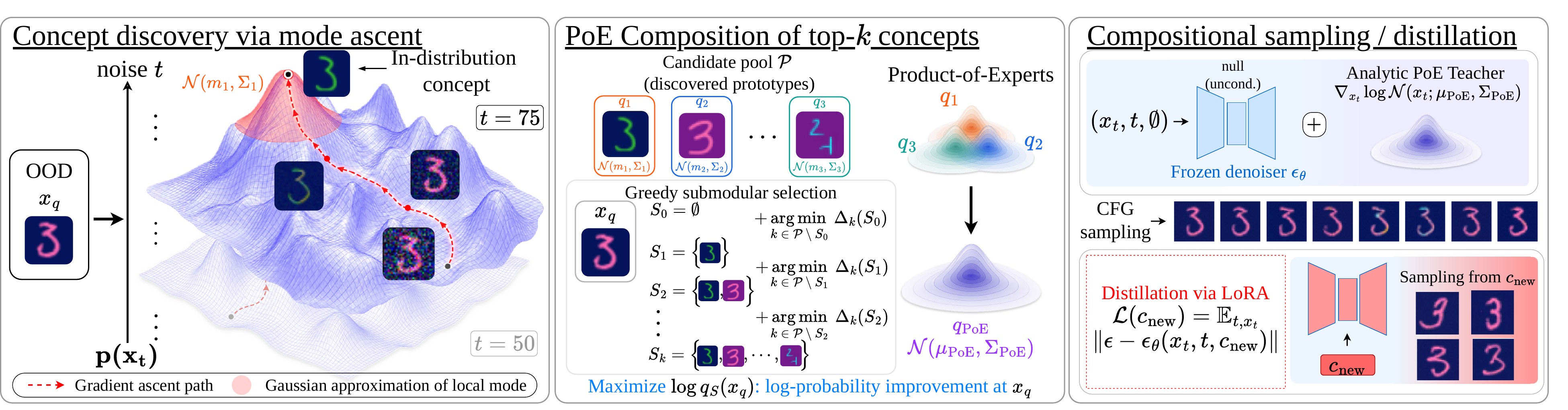}
    % \caption{An overall illustration of our concept discovery for compositional generalization with DDPM framework decipted in \Cref{sec:method}.}
    \caption{Overview of our DDPM-based concept discovery framework for compositional generalization, as described in \Cref{sec:method}.}
    \label{fig:placeholder}
\end{figure}

We study the problem of representing a query datum $x_q$ as a composition of concept prototypes discovered from a learned base distribution. A key feature of our setting is that no predefined primitives or underlying concepts are assumed to be given. Instead, the concept prototypes are recovered from the learned distribution itself, so the method requires neither a concept library nor per-concept supervision. Once composed, the concept prototypes define an analytic product-of-experts teacher model whose score is injected into a classifier-free, compositional diffusion sampler \cite{ho2022classifier,liu2022compositional}.
The remainder of this section develops the method in three stages. Section~\ref{sec:hierarchy} introduces the notion of a concept prototype as a mode of the noisy marginals $p_t$, locates such modes by gradient ascent on the unconditional score, and shows that modes are clean data and can therefore be approximated by a local Gaussian in $x_0$-space. Building on this, Section~\ref{sec:poe} combines the discovered concept prototypes into a product-of-experts whose score is available in closed form. Finally, Section~\ref{sec:sampling} plugs this score into classifier-free compositional sampling \cite{ho2022classifier,liu2022compositional} 
% and describes an optional distillation step that turns the discovered composition into a class conditioning token paired with a low-rank adaptation of the pretrained UNet.
and describes how we distill the discovered composition into the base model with LoRA~\citep{hu2022lora}.
We refer readers to \Cref{fig:placeholder} for an overview of the framework.

% \paragraph{Notation.}
% We work in the discrete-time diffusion setting \cite{sohl2015deep,ho2020denoising}. Let $x_0 \in \mathbb{R}^d$ denote a noise-free datum with distribution $p_{\text{data}}$, and let
% \begin{equation}
% x_t = \sqrt{\bar\alpha_t}\,x_0 + \sqrt{1-\bar\alpha_t}\,\varepsilon, \qquad \varepsilon\sim\mathcal{N}(0,I),\ t\in\{1,\ldots,T\},
% \label{eq:forward}
% \end{equation}
% be the forward-diffused state under schedule $\{\bar\alpha_t\}$. The noisy marginal is $p_t(x_t) = \int p_{\text{data}}(x_0)\,\mathcal{N}(x_t;\sqrt{\bar\alpha_t}x_0,(1-\bar\alpha_t)I)\,dx_0$, with score $\nabla_{x_t}\log p_t$ \cite{song2019generative,song2021score}. The network approximates this score through the noise-prediction head $\varepsilon_\theta(x_t,t,c)$, and the two are related by
% \begin{equation}
% s_\theta(x_t,t,c) = -\varepsilon_\theta(x_t,t,c)\big/\sqrt{1-\bar\alpha_t}.
% \label{eq:score-noise}
% \end{equation}
% We write $c$ for conditioning drawn from the trained set $\mathcal{C}=\{c_1,\ldots,c_N\}$, and $\emptyset$ for the null token. The unseen target concept is $c_*$. Discovered prototypes are indexed by $k=1,\ldots,K$ and sit at noise levels $t_k$. Each prototype carries an $x_t$-space Laplace covariance $\Sigma_k^{(t_k)}$ at its mode $x_t^{*,k}$ and is pulled back to an $x_0$-space Gaussian $q_k(x_0)=\mathcal{N}(x_0;m_k,\Sigma_k)$, with $m_k = x_t^{*,k}/\sqrt{\bar\alpha_{t_k}}$. Their weighted product of experts is $q_T$, and $c_{\text{new}}\in\mathbb{R}^{d_c}$ is the learned embedding that represents $c_*$ at inference.
\paragraph{Notation.}
We work in the discrete-time diffusion setting~\cite{sohl2015deep,ho2020denoising}. Let $x_0 \in \mathbb{R}^d$ denote a noise-free datum with distribution $p_{\mathrm{data}}$, and let
\begin{equation}
x_t = \sqrt{\bar\alpha_t}\,x_0 + \sqrt{1-\bar\alpha_t}\,\varepsilon, \qquad \varepsilon\sim\mathcal{N}(0,I),\quad t\in\{1,\ldots,T\},
\label{eq:forward}
\end{equation}
be the forward-diffused state under schedule $\{\bar\alpha_t\}$. The noisy marginal is
\(p_t(x_t) = \int p_{\mathrm{data}}(x_0)\,\mathcal{N}\!\left(x_t;\sqrt{\bar\alpha_t}x_0,(1-\bar\alpha_t)I\right)\,dx_0,
\)
with score $\nabla_{x_t}\log p_t(x_t)$~\cite{song2019generative,song2021score}. The network approximates this score through the noise-prediction head $\varepsilon_\theta(x_t,t,c)$, and the two are related by
\begin{equation}
s_\theta(x_t,t,c) = -\varepsilon_\theta(x_t,t,c)\big/\sqrt{1-\bar\alpha_t}.
\label{eq:score-noise}
\end{equation}
We write $c$ for conditioning drawn from the trained set $\mathcal{C}=\{c_1,\ldots,c_N\}$ and $\emptyset$ for the null token. The unseen target concept is $c_*$. Candidate modes are indexed by $j=1,\ldots,M$ and may come from different noise levels $t_j$. Each candidate has an $x_t$-space mode $x_t^{*,j}$, a Laplace covariance $\Sigma_j^{(t_j)}$ at that mode, and an $x_0$-space Gaussian expert $q_j(x_0)=\mathcal{N}(x_0;m_j,\Sigma_j)$ with $m_j=x_t^{*,j}/\sqrt{\bar\alpha_{t_j}}$. Greedy selection keeps $K$ concept prototypes, indexed by a selected set $S_K\subseteq\{1,\ldots,M\}$. Their weighted product-of-experts teacher is denoted $q_{\mathrm{T}}$, and $c_{\mathrm{new}}\in\mathbb{R}^{d_c}$ is the learned embedding used when this PoE teacher model is distilled into the base model using LoRA \cite{hu2022lora}.

\subsection{Concept discovery via mode ascent}
\label{sec:hierarchy}
A diffusion model admits a deep hierarchical latent-variable interpretation in which each $x_t$ is a latent whose prior is shaped by the learned score \cite{sohl2015deep,ho2020denoising,kingma2021variational,sonderby2016ladder,vahdat2020nvae}. 
A line of recent work examines how this hierarchy expresses itself across the noise schedule. Sclocchi et al. \cite{sclocchi2025phase} identify a sharp phase transition at an intermediate noise level, beyond which class identity is lost while low-level features persist, and argue that this behavior reflects a hierarchical, compositional structure in natural data. \begin{wrapfigure}{r}{0.48\linewidth}
    \centering
    \vspace{-5px}
    \includegraphics[width=\linewidth]{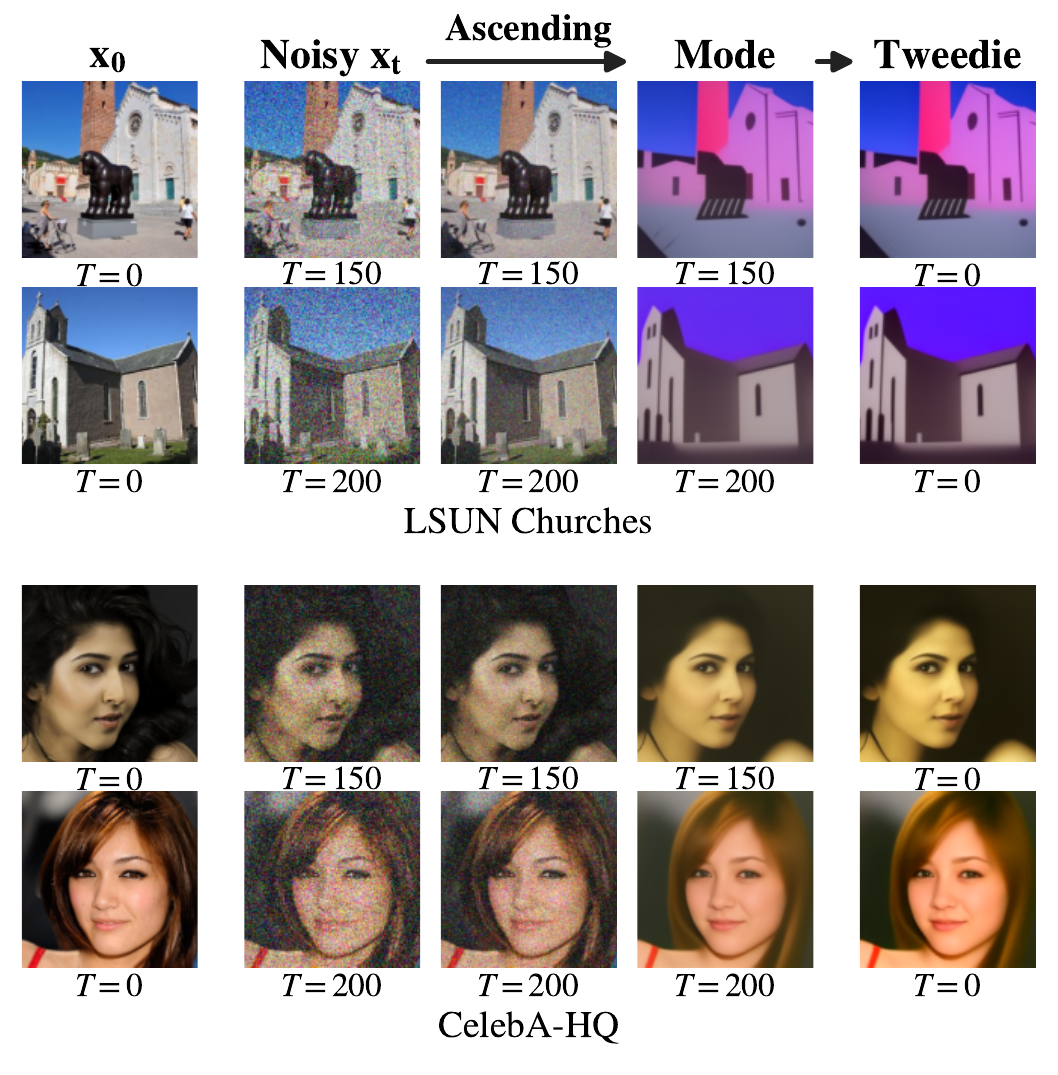}
    \vspace{-15px}\caption{Found local modes in pretrained DDPMs on LSUN Churches and CelebA-HQ.}
    \label{fig:mode-illu}
\end{wrapfigure}
% \hspace{-0.1cm}Complementary studies observe that the score carries coarse-to-fine information as $t$ decreases, so that large-$t$ regions govern high-level structure and small-$t$ regions refine details \cite{sclocchi2025phase,wang2025neuripsdeep}. 
\hspace{-0.1cm}Empirically, we observe a related hierarchy in the modes of the learned noisy marginals.
As the noise level increases, fine instance-level modes progressively merge into coarser concept prototypes, from which clear object-level classes emerge. See Appendix~\ref{app:fmnist}. 
Taken together, these results motivate treating the modes of $p_t$ at intermediate $t$ as a hierarchy of concept prototypes that the pretrained score function already encodes, to be recovered from the query at test time.

% By a \emph{mode} of $p_t$ we mean a local maximum of the noisy marginal, i.e., a solution of
% \begin{equation}
% x_t^* \in \operatornamewithlimits{arg\,local\,max}_{x_t}\ p_t(x_t).
% \label{eq:mode-def}
% \end{equation}
% Computing \Cref{eq:mode-def} in closed form is intractable because $p_t$ itself is available only implicitly through its score. We therefore locate modes by gradient ascent on the log-density,
% \begin{equation}
% x_t^{(i+1)} = x_t^{(i)} + \eta\,s_\theta\!\left(x_t^{(i)},\,t,\,\emptyset\right),
% \label{eq:mode-ascent}
% \end{equation}
% initialized at a noised copy of the query, $x_t^{(0)} \sim q(x_t\mid x_q)$. Using the unconditional score $s_\theta(\cdot,t,\emptyset)$ ensures that ascent explores the hierarchy of the full marginal $p_t$ rather than the conditional manifold of any trained class in $\mathcal{C}$, and in practice we replace the plain gradient step with an Adam update \cite{kingma2015adam}.
By a \emph{mode} of $p_t$ we mean a local maximum of the noisy marginal, i.e., a solution of
\begin{equation}
x_t^* \in \operatornamewithlimits{arg\,local\,max}_{x_t}\ p_t(x_t).
\label{eq:mode-def}
\end{equation}
Computing \Cref{eq:mode-def} in closed form is intractable because $p_t$ itself is available only implicitly through its score. We therefore locate modes by gradient ascent on the log-density,
\begin{equation}
x_t^{(i+1)} = x_t^{(i)} + \eta\,s_\theta\!\left(x_t^{(i)},\,t,\,\emptyset\right),
\label{eq:mode-ascent}
\end{equation}
initialized from a noised copy of the query,
\begin{equation*}
x_t^{(0)}=\sqrt{\bar\alpha_t}\,x_q+\sqrt{1-\bar\alpha_t}\,\varepsilon,\qquad \varepsilon\sim\mathcal{N}(0,I).
\end{equation*}
Using the unconditional score $s_\theta(\cdot,t,\emptyset)$ ensures that ascent explores the hierarchy of the full marginal $p_t$ rather than the conditional manifold of any trained class in $\mathcal{C}$. In practice, we replace the plain gradient step with an Adam update~\cite{kingma2015adam}.

\paragraph{Modes on the rescaled data manifold.}
% A random draw from $p_t$ is visibly noisy by Eq.~\eqref{eq:forward}, yet a mode of $p_t$ is not (see \Cref{fig:mode-illu}). The reason follows from Tweedie's formula \cite{robbins1956empirical,miyasawa1961empirical,efron2011tweedie,vincent2011connection}, which expresses the expected clean version of $x_t$ under the posterior as
% \begin{equation}
% \hat x_0(x_t) := \mathbb{E}[x_0\mid x_t] = \frac{1}{\sqrt{\bar\alpha_t}}\Big(x_t + (1-\bar\alpha_t)\,\nabla_{x_t}\log p_t(x_t)\Big).
% \label{eq:tweedie-mean}
% \end{equation}
% At a mode the score vanishes, and Eq.~\eqref{eq:tweedie-mean} reduces to $\hat x_0(x_t^*) = x_t^*/\sqrt{\bar\alpha_t}$. On the other hand, by the Gaussian-convolution structure of $p_t$, every local mode $m$ of $p_{\text{data}}$ induces a local mode of $p_t$ at approximately $\sqrt{\bar\alpha_t}\,m$ whenever neighboring modes are well separated relative to the kernel bandwidth $\sqrt{1-\bar\alpha_t}$. Combining the two,
% \begin{equation}
% x_t^* \approx \sqrt{\bar\alpha_t}\,m, \qquad \hat x_0(x_t^*) \approx m \in \mathrm{supp}(p_{\text{data}}).
% \label{eq:mode-on-manifold}
% \end{equation}
% In other words, a random $x_t\sim p_t$ carries a genuine $\sqrt{1-\bar\alpha_t}\,\varepsilon$ noise component that the rescaling $x_t/\sqrt{\bar\alpha_t}$ cannot remove, whereas a mode $x_t^*$ sits exactly at the peak of the convolution kernel, has no residual noise component, and rescales to a clean datum. The outputs of Eq.~\eqref{eq:mode-ascent} are therefore clean prototypes at every intermediate $t$, rather than noisy samples.
A random draw from $p_t$ is visibly noisy by Eq.~\eqref{eq:forward}, yet a mode of $p_t$ is noise-free (see \Cref{fig:mode-illu}). 
The reason follows from Tweedie's formula~\cite{robbins1956empirical,miyasawa1961empirical,efron2011tweedie,vincent2011connection}, which expresses the posterior mean of the clean datum as
\begin{equation}
\hat x_0(x_t) := \mathbb{E}[x_0\mid x_t] = \frac{1}{\sqrt{\bar\alpha_t}}\Big(x_t + (1-\bar\alpha_t)\,\nabla_{x_t}\log p_t(x_t)\Big).
\label{eq:tweedie-mean}
\end{equation}
At a mode, the score vanishes, so Eq.~\eqref{eq:tweedie-mean} reduces to $\hat x_0(x_t^*)=x_t^*/\sqrt{\bar\alpha_t}$. Conversely, by the Gaussian-convolution structure of $p_t$, a local mode $m$ of $p_{\mathrm{data}}$ induces a nearby local mode of $p_t$ at approximately $\sqrt{\bar\alpha_t}\,m$ whenever neighboring modes are well separated relative to the kernel bandwidth $\sqrt{1-\bar\alpha_t}$. Combining the two gives
\begin{equation}
x_t^* \approx \sqrt{\bar\alpha_t}\,m, \qquad \hat x_0(x_t^*) \approx m \in \mathrm{supp}(p_{\mathrm{data}}).
\label{eq:mode-on-manifold}
\end{equation}
Thus, unlike a random $x_t\sim p_t$, which contains a residual $\sqrt{1-\bar\alpha_t}\,\varepsilon$ noise component that simple rescaling cannot remove, a mode $x_t^*$ lies at a local density peak whose noise term vanishes.
We therefore treat the outputs of Eq.~\eqref{eq:mode-ascent} as noise-free concept prototypes rather than noisy samples.

% \paragraph{Local Gaussian at a mode.}
% Having located candidate modes $\{x_t^{*,j}\}_{j=1}^{M}$ at possibly different noise levels $\{t_j\}_{j=1}^{M}$, we summarize each one as a local Gaussian expert for later composition. 
% Because each local Gaussian expert initially live in different noisy variables $x_{t_j}$, we therefore rescale them to $x_0$-space, obtaining experts $q_j(x_0)=\mathcal{N}(x_0;m_j,\Sigma_j)$. 
% This common representation gives all experts a shared random variable $x_0$ for product-of-experts composition and allows each prototype to be forward-noised to any timestep needed by the sampler or LoRA distillation loss. 
% In implementation, we use diagonal covariances for $\Sigma_j$ estimated with 4 Hutchinson finite-difference probes.
\paragraph{Local Gaussian at a mode.}
Having located candidate modes $\{x_t^{*,j}\}_{j=1}^{M}$ at possibly different noise levels $\{t_j\}_{j=1}^{M}$, we summarize each one as a local Gaussian expert for later composition. 
% Because these experts are initially defined over different noisy variables $x_{t_j}$, we rescale them to $x_0$-space, obtaining experts $q_j(x_0)=\mathcal{N}(x_0;m_j,\Sigma_j)$. 
Because these experts are initially defined over different noisy variables $x_{t_j}$, we rescale them to $x_0$-space, obtaining experts $q_j(x_0)=\mathcal{N}(x_0;m_j,\Sigma_j)$ with $m_j=x_t^{*,j}/\sqrt{\bar\alpha_{t_j}}$.
This common representation gives all experts the shared random variable $x_0$ required for product-of-experts composition, and allows each prototype to be forward-noised to any timestep needed by the sampler or LoRA distillation loss. 
In implementation, we use diagonal covariances for $\Sigma_j$, estimated with four Hutchinson finite-difference probes.

\subsection{Product-of-experts composition}
\label{sec:poe}

% Let $\mathcal{P}=\{q_k(x_0)=\mathcal{N}(x_0;m_k,\Sigma_k)\}_{k=1}^{K}$ denote the candidate pool of Gaussian prototypes recovered by mode ascent and pulled back to $x_0$-space. Each discovered prototype is now an expert over a common random variable, so the natural way to combine them into a single distribution for the out-of-distribution concept is a weighted product of experts \cite{hinton2002training}. We use per-dimensional weights $w_k\in[0,1]^d$ under the diagonal Gaussian approximation, so that different prototypes can dominate different coordinates while ambiguous coordinates remain softly mixed. The product remains Gaussian, $q_T = \mathcal{N}(\mu_T,\Sigma_T)$, with closed-form parameters
% \begin{equation}
% q_T(x_0) \propto \prod_k q_k(x_0)^{w_k}, \quad
% \Sigma_T^{-1} = \sum_k \operatorname{Diag}(w_k)\,\Sigma_k^{-1}, \quad
% \mu_T = \Sigma_T \sum_k \operatorname{Diag}(w_k)\,\Sigma_k^{-1}m_k .
% \label{eq:poe-gaussian}
% \end{equation}
% Here the exponentiation is coordinate-wise, consistent with the diagonal covariance model. Combining experts in this way sharpens coordinates on which selected prototypes agree while preserving uncertainty where the evidence is mixed, which is the desired behavior when a novel concept is assembled from features supported by different prototypes \cite{du2019implicit,du2020compositional,liu2022compositional,du2023reduce}.
Let $\mathcal{P}=\{q_j(x_0)=\mathcal{N}(x_0;m_j,\Sigma_j)\}_{j=1}^{M}$ denote the candidate pool of Gaussian concept prototypes discovered by mode ascent and pulled back to $x_0$-space.
Each candidate is now an expert over a common random variable, so the natural way to combine selected prototypes into a distribution for the out-of-distribution concept is a weighted product of experts~\cite{hinton2002training}.
We use per-dimension weights $w_j\in[0,1]^d$ under the diagonal Gaussian approximation, allowing different prototypes to dominate different dimension while ambiguous dimension remain softly mixed. 
For a selected set $S_K$, the product remains Gaussian, $q_{\mathrm{T}}=\mathcal{N}(\mu_{\mathrm{T}},\Sigma_{\mathrm{T}})$, with closed-form parameters
\begin{equation}
q_{\mathrm{T}}(x_0) \propto \prod_{j\in S_K} q_j(x_0)^{w_j}, \quad
\Sigma_{\mathrm{T}}^{-1} = \sum_{j\in S_K} \operatorname{Diag}(w_j)\,\Sigma_j^{-1}, \quad
\mu_{\mathrm{T}} = \Sigma_{\mathrm{T}} \sum_{j\in S_K} \operatorname{Diag}(w_j)\,\Sigma_j^{-1}m_j .
\label{eq:poe-gaussian}
\end{equation}
Here exponentiation is dimension-wise, consistent with the diagonal covariance model. 
Combining experts in this way sharpens dimension on which selected prototypes agree while preserving uncertainty where the evidence is mixed, which is the desired behavior when a novel concept is assembled from features supported by different concept prototypes~\cite{du2019implicit,du2020compositional,liu2022compositional,du2023reduce}.

We select the prototype set $S_K\subseteq\{1,\ldots,M\}$ by greedy maximization of a per-dimension coverage objective. Write
\begin{equation*}
\ell_{j,r}(x_q) := \log\mathcal{N}\!\left(x_{q,r};\,m_{j,r},\,\sigma_{j,r}^2\right), \qquad r\in\{1,\ldots,d\},
\end{equation*}
where $\Sigma_j=\operatorname{Diag}(\sigma_{j,1}^2,\ldots,\sigma_{j,d}^2)$. This is the contribution of prototype $j$ to the log-likelihood of $x_q$ at dimension $r$. Under the diagonal PoE of Eq.~\eqref{eq:poe-gaussian}, the log-likelihood decomposes per dimension. We therefore use the coverage surrogate
\begin{equation}
F(S) \;:=\; \sum_{r=1}^{d}\, \max_{j \in S}\, \ell_{j,r}(x_q),
\label{eq:submodular-F}
\end{equation}
which is monotone and submodular in $S$~\cite{nemhauser1978analysis,krause2014submodular}.
We initialize with the best singleton, $S_1=\{\operatorname*{arg\,max}_{j}F(\{j\})\}$, and then grow the set by adding the candidate with the largest marginal gain:
\begin{equation}
j_{i+1} \;=\; \operatornamewithlimits{arg\,max}_{j\,\in\,\{1,\ldots,M\}\setminus S_i}\ \Delta_j(S_i), \qquad
\Delta_j(S_i) \;:=\; F(S_i \cup \{j\}) - F(S_i), \qquad
S_{i+1}=S_i\cup\{j_{i+1}\},
\label{eq:greedy}
\end{equation}
for $i=1,\ldots,K-1$. After selection, the per-dimension composition weight of each selected prototype is the temperature-controlled softmax of its per-dimension log-likelihood over the selected set,
\begin{equation}
w_j(r) \;=\; \frac{\exp\!\big(\ell_{j,r}(x_q)/\tau\big)}{\sum_{j' \in S_K} \exp\!\big(\ell_{j',r}(x_q)/\tau\big)}, \qquad j\in S_K,
\label{eq:per-dim-weights}
\end{equation}
so that dimensions where one prototype dominates concentrate weight on that prototype, while ambiguous dimensions retain a smoother mixture. 

A central consequence of the construction is that $q_{\mathrm{T}}$ is Gaussian in clean data space, and its score is analytic. The same is true of its forward-diffused counterpart
\begin{equation*}
q_{\mathrm{T}}^{(t)}(x_t)=\mathcal{N}\!\left(x_t;\sqrt{\bar\alpha_t}\mu_{\mathrm{T}},\,\bar\alpha_t\Sigma_{\mathrm{T}}+(1-\bar\alpha_t)I\right),
\end{equation*}
whose score is
\begin{equation}
\nabla_{x_t}\log q_{\mathrm{T}}^{(t)}(x_t) = -\big[\bar\alpha_t\Sigma_{\mathrm{T}} + (1-\bar\alpha_t)I\big]^{-1}\big(x_t - \sqrt{\bar\alpha_t}\,\mu_{\mathrm{T}}\big).
\label{eq:qT-score}
\end{equation}

\subsection{Compositional sampling and distillation}
\label{sec:sampling}

% Composable diffusion combines multiple concepts by adding their conditional score contributions, where each trained concept corresponds to a learned conditional score $\nabla_{x_t}\log p_t(x_t\mid c_k)$ \cite{liu2022compositional}. Our discovered concept is not a trained conditioning token or class label. Instead, the selected prototypes are composed into the analytic PoE teacher $q_T$, so the score of the discovered conditional distribution is available in closed form as the score of its noisy marginal $q_T^{(t)}$. Converting Eq.~\eqref{eq:qT-score} to the noise-prediction parameterization via Eq.~\eqref{eq:score-noise} gives
% \begin{equation}
% \varepsilon_{q_T}(x_t,t) = \sqrt{1-\bar\alpha_t}\,\big[\bar\alpha_t\Sigma_T + (1-\bar\alpha_t)I\big]^{-1}\big(x_t - \sqrt{\bar\alpha_t}\,\mu_T\big).
% \label{eq:eps-qT}
% \end{equation}
% We then use this analytic prediction as the conditional branch in classifier-free guidance \cite{ho2022classifier}:
% \begin{equation}
% \hat\varepsilon(x_t,t;c_*) =
% \varepsilon_\theta(x_t,t,\emptyset)
% +
% w_*(t)\Big(
% \varepsilon_{q_T}(x_t,t)
% -
% \varepsilon_\theta(x_t,t,\emptyset)
% \Big).
% \label{eq:comp-plusqT}
% \end{equation}
% Thus Eq.~\eqref{eq:comp-plusqT} samples the discovered concept by contrasting the analytic PoE prediction against the null prediction, without requiring a trained conditional token for $c_*$. Equations~\eqref{eq:pullback}--\eqref{eq:eps-qT} therefore form a closed pipeline from the query $x_q$ to a guidance direction that the CFG/compositional sampler can consume directly.
Composable diffusion combines multiple concepts by adding their conditional score contributions, where each trained concept corresponds to a learned conditional score $\nabla_{x_t}\log p_t(x_t\mid c_k)$~\cite{liu2022compositional}. Our discovered concept is not a trained conditioning token or class label. Instead, the selected concept prototypes define the analytic PoE teacher model $q_{\mathrm{T}}$, so the score of the discovered conditional distribution is available in closed form as the score of its noisy marginal $q_{\mathrm{T}}^{(t)}$. Converting Eq.~\eqref{eq:qT-score} to the noise-prediction parameterization via Eq.~\eqref{eq:score-noise} gives
\begin{equation}
\varepsilon_{q_{\mathrm{T}}}(x_t,t) = \sqrt{1-\bar\alpha_t}\,\big[\bar\alpha_t\Sigma_{\mathrm{T}} + (1-\bar\alpha_t)I\big]^{-1}\big(x_t - \sqrt{\bar\alpha_t}\,\mu_{\mathrm{T}}\big).
\label{eq:eps-qT}
\end{equation}
We then use this analytic prediction as the conditional branch in classifier-free guidance~\cite{ho2022classifier}:
\begin{equation}
\hat\varepsilon(x_t,t;c_*) =
\varepsilon_\theta(x_t,t,\emptyset)
+
w_*(t)\Big(
\varepsilon_{q_{\mathrm{T}}}(x_t,t)
-
\varepsilon_\theta(x_t,t,\emptyset)
\Big).
\label{eq:comp-plusqT}
\end{equation}
Thus Eq.~\eqref{eq:comp-plusqT} samples the discovered concept by contrasting the analytic PoE prediction against the null prediction, without requiring a trained conditional token for $c_*$. Equations~\eqref{eq:mode-ascent}--\eqref{eq:eps-qT} therefore form a closed framework from the query $x_q$ to a guidance direction that the CFG/compositional sampler can consume directly.

% \paragraph{Variance-aware guidance schedule.}
% The residual $\varepsilon_{q_T}(x_t,t)-\varepsilon_\theta(x_t,t,\emptyset)$ inherits the precision factor $\Sigma_t^{-1}$ from Eq.~\eqref{eq:eps-qT}, where $\Sigma_t := \bar\alpha_t\Sigma_T+(1-\bar\alpha_t)I$ is the diagonal covariance of the noisy PoE marginal. A tight teacher with small $\Sigma_T$ can therefore produce an outsized pull at small $t$, while a diffuse teacher can produce a faint one. To make the guidance strength of $c_*$ comparable across queries irrespective of prototype geometry, we set
% \begin{equation}
% w_*(t) \;=\; \min\!\Big(w_0 \cdot \tfrac{1}{d}\,\mathrm{tr}(\Sigma_t),\; w_{\max}\Big),
% \label{eq:variance-guidance}
% \end{equation}
% which approximately cancels the implicit $\Sigma_t^{-1}$ scaling, while the cap $w_{\max}$ prevents blow-up at large $t$ where $\Sigma_t \to I$. 
% % The base scale $w_0$ and cap $w_{\max}$ are dataset-dependent and are reported with the corresponding experimental setup.
\paragraph{Variance-aware guidance schedule.}
The residual $\varepsilon_{q_{\mathrm{T}}}(x_t,t)-\varepsilon_\theta(x_t,t,\emptyset)$ inherits the precision factor $\Sigma_t^{-1}$ from Eq.~\eqref{eq:eps-qT}, where $\Sigma_t:=\bar\alpha_t\Sigma_{\mathrm{T}}+(1-\bar\alpha_t)I$ is the diagonal covariance of the noisy PoE marginal. A concentrated PoE distribution with small $\Sigma_{\mathrm{T}}$ can therefore produce an outsized pull at small $t$, while a diffuse PoE can produce a faint one. To make the guidance strength of $c_*$ comparable across queries irrespective of prototype geometry, we set
\begin{equation}
w_*(t) \;=\; \min\!\Big(w_0 \cdot \tfrac{1}{d}\,\mathrm{tr}(\Sigma_t),\; w_{\max}\Big),
\label{eq:variance-guidance}
\end{equation}
which approximately cancels the implicit $\Sigma_t^{-1}$ scaling, while the cap $w_{\max}$ prevents blow-up at large $t$ where $\Sigma_t\to I$.

% \paragraph{Distillation with LoRA.}
% We can further distill the PoE teacher into the base model by jointly learning a new class embedding $c_{\text{new}}$ and a LoRA adapter \cite{hu2022lora} attached to the frozen UNet. Let $\mathcal{D}_T = \{x_0^{(i)}\}_{i=1}^{N_{\text{pool}}}$ be a fixed pool obtained from the sampler of Eq.~\eqref{eq:comp-plusqT} with the schedule of Eq.~\eqref{eq:variance-guidance}, and let $\varepsilon_{\theta+\theta'}$ denote the noise predictor with the trainable parameters $\theta' = \{\theta_{\text{LoRA}},\,c_{\text{new}}\}$. We optimize the standard noise-prediction loss
% \begin{equation}
% \mathcal{L}(\theta') = \mathbb{E}_{x_0\sim \mathcal{D}_T,\,t,\,\varepsilon}\,\Big\|\varepsilon - \varepsilon_{\theta+\theta'}\!\big(\sqrt{\bar\alpha_t}\,x_0 + \sqrt{1-\bar\alpha_t}\,\varepsilon,\ t,\ c_{\text{new}}\big)\Big\|^2,
% \label{eq:lora-distill-loss}
% \end{equation}
% with CFG-style dropout on $c_{\text{new}}$ during training so that the unconditional branch remains usable at inference. Once trained, the OOD concept $c_*$ is absorbed into the base model and can be generated by standard classifier-free sampling \cite{ho2022classifier} conditioned on $c_{\text{new}}$, replacing the analytic predictor of Eq.~\eqref{eq:comp-plusqT}. Related single-concept personalization formulations include textual inversion \cite{gal2022image}, DreamBooth \cite{ruiz2023dreambooth}, and Custom Diffusion \cite{kumari2023multi}; our distillation differs in that the optimization target is the analytic PoE teacher $q_T$ rather than user-provided exemplars.
\paragraph{Distillation with LoRA.}
We can further distill the PoE teacher model into the base model by jointly learning a new class embedding $c_{\mathrm{new}}$ and a LoRA adapter~\cite{hu2022lora} attached to the frozen denoiser network. Let $\mathcal{D}_{\mathrm{T}}=\{x_0^{(i)}\}_{i=1}^{N_{\mathrm{pool}}}$ be a fixed pool obtained from the sampler of Eq.~\eqref{eq:comp-plusqT} with the schedule of Eq.~\eqref{eq:variance-guidance}, and let $\varepsilon_{\theta+\theta'}$ denote the denoiser with trainable parameters $\theta'=\{\theta_{\mathrm{LoRA}},c_{\mathrm{new}}\}$. We optimize the standard noise-prediction loss
\begin{equation}
\mathcal{L}(\theta') = \mathbb{E}_{x_0\sim\mathcal{D}_{\mathrm{T}},\,t,\,\varepsilon}\,\Big\|\varepsilon - \varepsilon_{\theta+\theta'}\!\big(\sqrt{\bar\alpha_t}\,x_0 + \sqrt{1-\bar\alpha_t}\,\varepsilon,\ t,\ c_{\mathrm{new}}\big)\Big\|^2,
\label{eq:lora-distill-loss}
\end{equation}
with CFG-style dropout on $c_{\mathrm{new}}$ during training so that the unconditional branch remains usable at inference. Once trained, the OOD concept $c_*$ can be generated by standard classifier-free sampling~\cite{ho2022classifier} conditioned on $c_{\mathrm{new}}$, replacing the analytic predictor of Eq.~\eqref{eq:comp-plusqT}. Unlike single-concept personalization formulations include textual inversion~\cite{gal2022image}, DreamBooth~\cite{ruiz2023dreambooth}, and Custom Diffusion~\cite{kumari2023multi}, our distillation differs in that the optimization target is the analytic PoE teacher model $q_{\mathrm{T}}$ rather than user-provided exemplars.

% Taken together, these steps close the concept-discovery gap left by prior compositional-diffusion work \cite{du2020compositional,liu2022compositional,du2023reduce}, which assumes a given library of primitive concepts, each already represented by a trained score.
% Mode ascent on the pretrained unconditional score recovers prototypes from a single out-of-distribution query, a product of experts composes them into an analytic teacher $q_T$ with closed-form score, and the teacher slots directly into the classifier-free compositional sampler.
% The optional distillation step then turns the result into a single embedding plus a low-rank UNet adaptation that promote $c_*$ to a first-class citizen of the model's conditioning interface, so that concept discovery and compositional generation become end-to-end from the query alone.
Taken together, these steps close the concept discovery gap left by prior compositional-diffusion work~\cite{du2020compositional,liu2022compositional,du2023reduce}, which assumes a given library of primitive concepts, each already represented by a trained score. Mode ascent on the pretrained unconditional score recovers concept prototypes from a single out-of-distribution query, a product-of-experts composes them into an analytic distribution $q_{\mathrm{T}}$ with closed-form score, and its score slots directly into the classifier-free compositional sampler. The optional LoRA distillation step then absorbs the discovered composition into the model's conditioning interface, making concept discovery and compositional generation possible from the query alone.

% =============================================================
% Main-text fragment: Experiments section
% =============================================================
\section{Test-Time Diffusion Compositional Generalization on OOD Query}
\label{sec:exp}

\subsection{Compositional benchmark}
\label{sec:exp:benchmark}

Test-time compositional generalization asks whether, given a single OOD query $x_q$ depicting a held-out combination of primitive attributes, a pretrained diffusion model can sample a distribution of images of that combination, without access to any prebuilt concept library.
We evaluate this on the following two compositional benchmarks:
% , that share the same construction recipe: factorize each image into primitive attributes, enumerate the Cartesian product as compositional slots, and partition the slots into a seen set used to train a class-conditional diffusion backbone with classifier-free guidance \cite{ho2022classifier} and an unseen set accessible only at test time.

\paragraph{Color-MNIST.}
A $32{\times}32$ RGB rendering of MNIST with three primitive factors: digit identity ($10$ values, $0$--$9$), digit color ($4$ values from a muted high-contrast palette: yellow, green, cyan, pink), and background color ($4$ values: deep red, navy, dark purple, dark brown). The Cartesian product yields $10 \times 4 \times 4 = 160$ compositional slots, of which $120$ are seen by the backbone during training and $40$ are held out as OOD classes.

\paragraph{CelebA (compositional).}
We construct a compositional benchmark from binary CelebA \cite{liu2015deep} face attributes such as hair color, smiling, beard, and gender. Each unique realized combination of $14$ such binary attributes defines a compositional class. Of the $74$ realized combinations used in this experiment, $51$ form the seen split for training and $23$ are held out as OOD classes. Images are rendered at $128{\times}128$.

We provide additional details on both datasets in Appendix~\ref{app:benchmark} and diffusion training in Appendix~\ref{app:backbone}.
% For every OOD class, we draw a query $x_q$ and run our framework (concept discovery, PoE composition, CFG sampling, and optional distillation) to produce one generated image; we repeat this for $10$ distinct queries $\{x_q^{(i)}\}_{i=1}^{10}$ per OOD class to obtain a $10$-image evaluation set. The remaining test-set images of the same class, with the $10$ query indices excluded, form a leak-free real reference bank used for evaluation. Full attribute lists, per-class image counts, and the data-preparation pipeline are deferred to Appendix~\ref{app:benchmark}.

\subsection{Hypothesis, baselines, and metrics}
\label{sec:exp:baselines}

\paragraph{Hypothesis.}
A single OOD query of unseen compositions $x_q$ contains enough information for the pretrained diffusion model to expose a multi-mode neighborhood at different noising timesteps that approximates the held-out compositional class. 
Concretely, the discovered prototypes $\{m_j,\Sigma_j\}$ aggregated into the PoE teacher $q_T$ of Eq.~\eqref{eq:poe-gaussian} should support classifier-free sampling that both (i) reconstructs the OOD concept around the query and (ii) generalizes beyond the specific query to other instances of the same OOD class. Distillation of the PoE teacher into a learned $(c_{\text{new}},\,\theta_{\text{LoRA}})$ pair (Eq.~\eqref{eq:lora-distill-loss}) further absorbs the discovered concept onto the base model's score manifold, which we expect to yield a richer and more on-distribution conditional for sampling.

\paragraph{Protocol.}
For each OOD class, we run the full framework of Section~\ref{sec:method} (mode ascent, PoE composition, variance-aware CFG sampling, and distillation) independently on each of the $100$ queries $x_q^{(i)}$. We evaluate two configurations of our method:
\begin{itemize}
  \item \textbf{PoE analytic CFG.} The classifier-free sampler of Eq.~\eqref{eq:comp-plusqT} that uses the analytic PoE score $\varepsilon_{q_T}$ from Eq.~\eqref{eq:eps-qT} as the conditional branch, with the variance-aware schedule of Eq.~\eqref{eq:variance-guidance}.
  % \item \textbf{LoRA distillation.} The PoE analytic-CFG sampling pool of $256$ images is distilled into the base model by jointly learning a LoRA adapter on the frozen UNet and a new class token $c_{\text{new}}$ (Eq.~\eqref{eq:lora-distill-loss}). Inference uses standard classifier-free sampling on the LoRA-adapted UNet conditioned on $c_{\text{new}}$.
  \item \textbf{LoRA.} We distill $256$ PoE-generated samples into the base model by training a LoRA adapter on the frozen UNet together with a new class token $c_{\text{new}}$ (Eq.~\eqref{eq:lora-distill-loss}). At inference, we sample from the LoRA-adapted model conditioned on $c_{\text{new}}$.
\end{itemize}

\paragraph{Top-$k$ trained classes baseline.}
For each query, we score every trained class $c \in \mathcal{C}$ by the average DDPM loss of $x_q$ under that conditional,
\[
\ell(c) := \mathbb{E}_{t \in \mathcal{T}_{\mathrm{eval}},\,\varepsilon}
\big\|\varepsilon - \varepsilon_\theta(\sqrt{\bar\alpha_t}x_q+\sqrt{1-\bar\alpha_t}\varepsilon,\,t,\,c)\big\|^2,
\]
and pick the $k$ smallest-loss classes $c_{(1)},\ldots,c_{(k)}$. The top-$1$ baseline replaces $\varepsilon_{q_T}(x_t,t)$ in Eq.~\eqref{eq:comp-plusqT} with $\varepsilon_\theta(x_t,t,c_{(1)})$. The top-$3$ baseline composes the three smallest-loss classes via the multi-concept composition of \cite{liu2022compositional} with weights $w_k = \mathrm{softmax}_k(-\ell(c_{(k)})/\tau_{\mathrm{tk}})$. Both test whether concept discovery from the pretrained diffusion model recovers richer information about the OOD composition than retrieving or compositing the nearest seen class tokens.

% \paragraph{Query-only ($q_{x_q}$) baseline.}
% The analytic PoE teacher model is replaced with a single-Gaussian teacher model $q_{x_q}(x_0) = \mathcal{N}(x_q,\,\sigma^2 I)$, in which we impose a small isotropic variance around $x_q$ to approximate a distribution from a single point. 
% The teacher model is routed through the same Eq.~\eqref{eq:comp-plusqT} sampler.
% This tests whether our PoE teacher model encodes richer compositional information than point-estimate. In contrast, the PoE teacher's covariance is derived from the local Hessian of the score field at each discovered mode and therefore encodes the score-implied uncertainty about the OOD concept rather than a fixed spread around the query.
\paragraph{Query-only ($q_{x_q}$) baseline.}
The analytic PoE teacher model is replaced with a single-query Gaussian teacher model
$q_{x_q}(x_0)=\mathcal{N}(x_q,\sigma^2 I)$, where a fixed small isotropic variance is placed around the query image $x_q$. 
This teacher model is routed through the same sampler as Eq.~\eqref{eq:comp-plusqT}. 
This baseline tests whether our PoE teacher model captures richer compositional structure than a distribution built only by spreading fixed variance around a single query. 
In contrast, the PoE teacher model combines multiple discovered concept prototypes and uses their score-implied local covariance, rather than a fixed isotropic spread around $x_q$.

% \paragraph{Metrics and reference sets.}
% For each OOD class, we compute the following metrics: FID \cite{heusel2017gans}, CLIP image--image cosine similarity \cite{radford2021learning}, precision and recall use the $k$-NN density estimator of \cite{kynkaanniemi2019improved} in Inception-V3 feature space from FID with $k{=}3$, and F1 is their harmonic mean.
% To separate \emph{faithfulness to the query} from \emph{generalization beyond it}, we compute every metric between samples generated from each compared approach (PoE, LoRA, top-$1/3$, and query-only) and two reference sets per OOD class:
\paragraph{Metrics and reference sets.}
For each OOD class, we evaluate generated samples using FID~\cite{heusel2017gans}, CLIP image--image cosine similarity~\cite{radford2021learning}, and precision/recall in Inception-V3 feature space using the $k$-NN density estimator \cite{kynkaanniemi2019improved} with $k{=}3$. 
We report F1 as the harmonic mean of precision and recall. 
To distinguish \emph{faithfulness to the query} from \emph{generalization beyond the query}, we compare samples from each method (PoE, LoRA, top-$1$, top-$3$, and query-only) against two reference sets for each OOD class:
\begin{itemize}
  \item \textbf{Faithfulness.} The $100$ query images $\{x_q^{(i)}\}$ themselves. Asks whether generations concentrate near the queries that drove concept discovery.
  % \item \textbf{Generalization.} $100$ other held-out images of the same OOD class drawn at random with the query indices excluded. Asks whether generations generalize beyond the specific seen queries.
  \item \textbf{Generalization.} A random set of $100$ other held-out images from the same OOD class, excluding the query. This measures whether generations extend beyond the specific queries.
\end{itemize}
% Each metric is computed per OOD class on the matched $10$-vs-$10$ sets and aggregated to mean $\pm$ standard error across the OOD-class population ($40$ Color-MNIST classes, $23$ CelebA holdout classes). 
% Full feature extractors, exact reference-set construction, and the small-$N$ caveats for FID are detailed in Appendix~\ref{app:eval}.
% We refer to Appendix \ref{app:hypers} for detials on concept discovery, sampling, and distillation configuration and hyperparameters and \ref{app:eval} for small-$N$ caveats for FID.
We refer readers to Appendix~\ref{app:hypers} for details on the concept discovery, sampling, and distillation configurations and hyperparameters, and to Appendix~\ref{app:eval} for small-$N$ caveats related to FID.

\subsection{Results and analysis}
\label{sec:exp:results}
\begin{figure}[ht]
    \centering
    \includegraphics[width=1\linewidth]{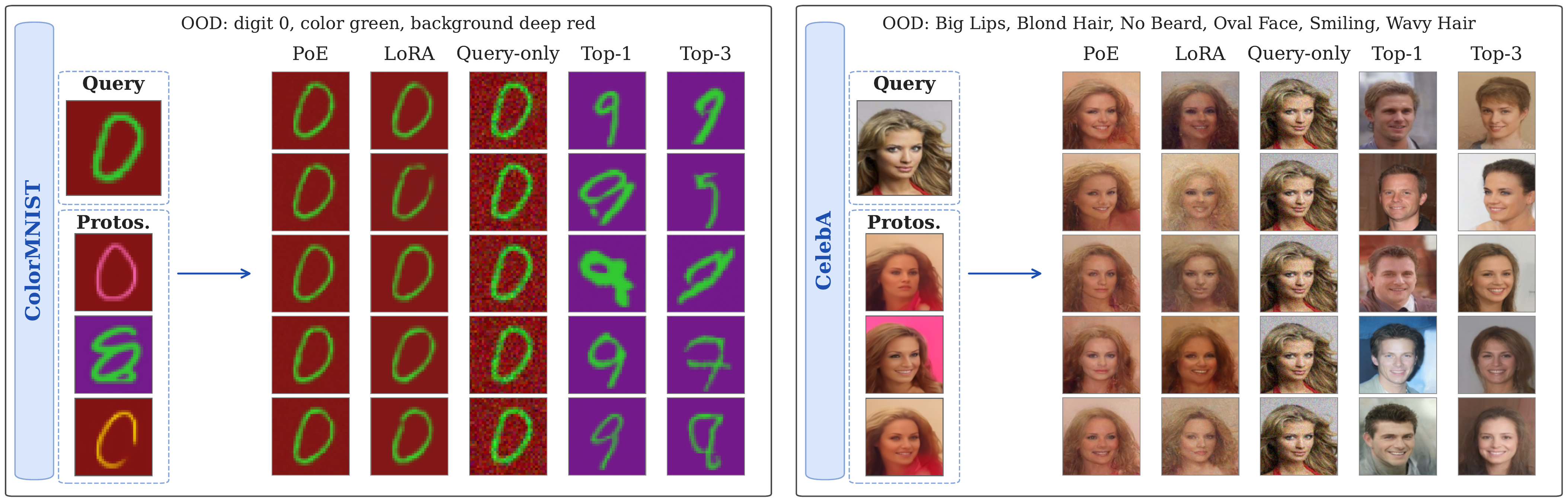}
    \caption{Examples of found prototypes, given an OOD query of unseen compositions, and generated samples from each of compared methods in ColorMNIST and CelebA dataset. We refer readers to Appendix~\ref{app:viz} for additional qualitative results.}
    \label{fig:dual-panel}
\end{figure}
\begin{table}[ht]
\centering
\caption{ColorMNIST: 40 OOD compositions ($\text{mean}_{\scriptscriptstyle\pm\text{SE}}$). \textbf{Bold}: best; \underline{underline}: 2nd.}
\vspace{0.5em}
\label{tab:cmnist_main}
\scriptsize
\setlength{\tabcolsep}{2.2pt}
\renewcommand{\arraystretch}{1.3}
\begin{tabular}{lrrrrrrrrrr}
\toprule
 & \multicolumn{2}{c}{FID $\downarrow$} & \multicolumn{2}{c}{CLIP $\uparrow$ (\%)} & \multicolumn{2}{c}{Precision $\uparrow$ (\%)} & \multicolumn{2}{c}{Recall $\uparrow$ (\%)} & \multicolumn{2}{c}{F1 $\uparrow$ (\%)} \\
\cmidrule(lr){2-3} \cmidrule(lr){4-5} \cmidrule(lr){6-7} \cmidrule(lr){8-9} \cmidrule(lr){10-11}
Method & \textit{Faithful.} & \textit{General.} & \textit{Faithful.} & \textit{General.} & \textit{Faithful.} & \textit{General.} & \textit{Faithful.} & \textit{General.} & \textit{Faithful.} & \textit{General.} \\
\midrule
Top-1 & $191.2_{\scriptscriptstyle\pm4.8}$ & $194.4_{\scriptscriptstyle\pm4.6}$ & $91.0_{\scriptscriptstyle\pm0.2}$ & $\underline{91.0}_{\scriptscriptstyle\pm0.2}$ & $9.5_{\scriptscriptstyle\pm3.4}$ & $5.3_{\scriptscriptstyle\pm2.2}$ & $42.2_{\scriptscriptstyle\pm6.6}$ & $38.8_{\scriptscriptstyle\pm6.2}$ & $6.2_{\scriptscriptstyle\pm2.2}$ & $6.5_{\scriptscriptstyle\pm2.6}$ \\
Top-3 & $178.5_{\scriptscriptstyle\pm5.4}$ & $181.0_{\scriptscriptstyle\pm5.3}$ & $90.2_{\scriptscriptstyle\pm0.2}$ & $90.3_{\scriptscriptstyle\pm0.2}$ & $11.0_{\scriptscriptstyle\pm3.3}$ & $15.8_{\scriptscriptstyle\pm3.5}$ & $30.0_{\scriptscriptstyle\pm5.2}$ & $30.5_{\scriptscriptstyle\pm5.5}$ & $5.1_{\scriptscriptstyle\pm1.6}$ & $5.3_{\scriptscriptstyle\pm1.7}$ \\
Query-only & $247.1_{\scriptscriptstyle\pm6.1}$ & $254.9_{\scriptscriptstyle\pm6.0}$ & $82.0_{\scriptscriptstyle\pm0.3}$ & $81.9_{\scriptscriptstyle\pm0.3}$ & $0.5_{\scriptscriptstyle\pm0.3}$ & $0.0_{\scriptscriptstyle\pm0.0}$ & $2.5_{\scriptscriptstyle\pm1.4}$ & $1.2_{\scriptscriptstyle\pm1.2}$ & $0.0_{\scriptscriptstyle\pm0.0}$ & $0.0_{\scriptscriptstyle\pm0.0}$ \\
\midrule
PoE & $\underline{81.8}_{\scriptscriptstyle\pm2.6}$ & $\underline{102.9}_{\scriptscriptstyle\pm3.4}$ & $\underline{91.2}_{\scriptscriptstyle\pm0.4}$ & $90.9_{\scriptscriptstyle\pm0.4}$ & $\underline{75.5}_{\scriptscriptstyle\pm3.3}$ & $\underline{67.0}_{\scriptscriptstyle\pm3.4}$ & $\underline{80.7}_{\scriptscriptstyle\pm3.3}$ & $\underline{69.8}_{\scriptscriptstyle\pm3.6}$ & $\underline{75.3}_{\scriptscriptstyle\pm3.1}$ & $\underline{63.0}_{\scriptscriptstyle\pm2.7}$ \\
\(+\)LoRA & $\mathbf{62.6}_{\scriptscriptstyle\pm2.5}$ & $\mathbf{84.7}_{\scriptscriptstyle\pm2.8}$ & $\mathbf{95.1}_{\scriptscriptstyle\pm0.1}$ & $\mathbf{94.8}_{\scriptscriptstyle\pm0.2}$ & $\mathbf{89.5}_{\scriptscriptstyle\pm2.0}$ & $\mathbf{82.2}_{\scriptscriptstyle\pm2.7}$ & $\mathbf{88.0}_{\scriptscriptstyle\pm2.8}$ & $\mathbf{78.8}_{\scriptscriptstyle\pm3.8}$ & $\mathbf{87.6}_{\scriptscriptstyle\pm2.1}$ & $\mathbf{76.3}_{\scriptscriptstyle\pm2.6}$ \\
\bottomrule
\end{tabular}
\end{table}
\begin{table}[ht]
\centering
\caption{CelebA: 23 OOD compositions ($\text{mean}_{\scriptscriptstyle\pm\text{SE}}$). \textbf{Bold}: best; \underline{underline}: 2nd.}
\vspace{0.5em}
\label{tab:celeba_main}
\scriptsize
\setlength{\tabcolsep}{2pt}
\renewcommand{\arraystretch}{1.3}
\begin{tabular}{lrrrrrrrrrr}
\toprule
 & \multicolumn{2}{c}{FID $\downarrow$} & \multicolumn{2}{c}{CLIP $\uparrow$ (\%)} & \multicolumn{2}{c}{Precision $\uparrow$ (\%)} & \multicolumn{2}{c}{Recall $\uparrow$ (\%)} & \multicolumn{2}{c}{F1 $\uparrow$ (\%)} \\
\cmidrule(lr){2-3} \cmidrule(lr){4-5} \cmidrule(lr){6-7} \cmidrule(lr){8-9} \cmidrule(lr){10-11}
Method & \textit{Faithful.} & \textit{General.} & \textit{Faithful.} & \textit{General.} & \textit{Faithful.} & \textit{General.} & \textit{Faithful.} & \textit{General.} & \textit{Faithful.} & \textit{General.} \\

\midrule
Top-1 & $193.6_{\scriptscriptstyle\pm2.3}$ & $199.5_{\scriptscriptstyle\pm2.6}$ & $60.4_{\scriptscriptstyle\pm0.4}$ & $60.2_{\scriptscriptstyle\pm0.4}$ & $39.4_{\scriptscriptstyle\pm4.3}$ & $43.1_{\scriptscriptstyle\pm4.5}$ & $1.4_{\scriptscriptstyle\pm0.6}$ & $1.0_{\scriptscriptstyle\pm0.5}$ & $1.9_{\scriptscriptstyle\pm0.8}$ & $1.5_{\scriptscriptstyle\pm0.7}$ \\
Top-3 & $179.0_{\scriptscriptstyle\pm2.4}$ & $191.9_{\scriptscriptstyle\pm2.7}$ & $61.6_{\scriptscriptstyle\pm0.4}$ & $61.3_{\scriptscriptstyle\pm0.4}$ & $45.3_{\scriptscriptstyle\pm5.4}$ & $48.9_{\scriptscriptstyle\pm5.2}$ & $0.4_{\scriptscriptstyle\pm0.3}$ & $0.7_{\scriptscriptstyle\pm0.4}$ & $0.6_{\scriptscriptstyle\pm0.3}$ & $1.0_{\scriptscriptstyle\pm0.6}$ \\
Query-only & $\mathbf{107.5}_{\scriptscriptstyle\pm1.4}$ & $\underline{147.5}_{\scriptscriptstyle\pm2.8}$ & $66.6_{\scriptscriptstyle\pm0.7}$ & $66.3_{\scriptscriptstyle\pm0.8}$ & $\mathbf{80.6}_{\scriptscriptstyle\pm2.3}$ & $48.2_{\scriptscriptstyle\pm3.1}$ & $\mathbf{53.6}_{\scriptscriptstyle\pm2.9}$ & $\underline{13.4}_{\scriptscriptstyle\pm1.3}$ & $\mathbf{64.0}_{\scriptscriptstyle\pm2.7}$ & $\underline{20.2}_{\scriptscriptstyle\pm1.7}$ \\
\midrule
PoE & $\underline{120.9}_{\scriptscriptstyle\pm3.1}$ & $\mathbf{140.3}_{\scriptscriptstyle\pm3.9}$ & $\mathbf{68.5}_{\scriptscriptstyle\pm0.8}$ & $\mathbf{68.4}_{\scriptscriptstyle\pm0.9}$ & $\underline{75.8}_{\scriptscriptstyle\pm1.4}$ & $\mathbf{74.5}_{\scriptscriptstyle\pm1.8}$ & $12.0_{\scriptscriptstyle\pm1.0}$ & $9.0_{\scriptscriptstyle\pm1.0}$ & $20.3_{\scriptscriptstyle\pm1.4}$ & $15.6_{\scriptscriptstyle\pm1.4}$ \\
\(+\)LoRA & $142.4_{\scriptscriptstyle\pm3.9}$ & $160.1_{\scriptscriptstyle\pm4.4}$ & $\underline{67.5}_{\scriptscriptstyle\pm0.7}$ & $\underline{67.3}_{\scriptscriptstyle\pm0.8}$ & $46.2_{\scriptscriptstyle\pm2.1}$ & $\underline{52.7}_{\scriptscriptstyle\pm3.0}$ & $\underline{19.9}_{\scriptscriptstyle\pm1.4}$ & $\mathbf{16.8}_{\scriptscriptstyle\pm1.2}$ & $\underline{26.9}_{\scriptscriptstyle\pm1.4}$ & $\mathbf{24.7}_{\scriptscriptstyle\pm1.6}$ \\
\bottomrule
\end{tabular}
\end{table}

 % compare two ways of consuming the discovered prototypes. PoE uses the modes directly through the analytic teacher $q_T$, while LoRA distils samples from this teacher back into the frozen backbone with a learned adapter and class token. The purpose of LoRA is therefore not to compete with the PoE sample pool, but to test whether the discovered OOD composition can be written back onto the model manifold. Under this reading, the key comparison is against Query only and the Top 1 and Top 3 trained class baselines.
Across both datasets in \Cref{tab:cmnist_main,tab:celeba_main}, PoE and LoRA are more faithful to $x_q$ and generalize better to other held out examples of the same OOD class.
On ColorMNIST, PoE and LoRA outperform Query-only and the top-$k$ baselines on both Faithfulness and Generalization, and LoRA further improves over PoE across the table.
Its F1 remains high from Faithfulness to Generalization ($87.6\%$ and $76.3\%$), suggesting that distillation promotes in manifold exploration rather than memorization of the query. 
The ColorMNIST panel of \Cref{fig:dual-panel} explains why the trained class baselines are weak proxies for an unseen composition. 
The Top 1 and Top 3 columns often match surface similarity while changing the digit identity, which is consistent with their low F1 values near $5\%$ to $7\%$.
This supports the view that richer information about an unseen composition lies beyond nearest trained class retrieval. 
The Query-only column gives the complementary failure mode. 
Sampling around a single fixed image produces salt and pepper texture because the covariance carries little information about the true class variance. 
By contrast, the discovered prototypes already preserve the correct digit, foreground color, and background color, and the PoE covariance uses the local score geometry around these modes to encode manifold spread.

% On CelebA, the same argument becomes sharper because the images contain greater intra-class diversity and higher pixel-level information density. 
% Although Query-only attains the strongest Faithfulness scores, including $92.6\%$ F1, its Generalization F1 drops to $47.4\%$. 
% This gap indicates that $q_{x_q}$ captures the data point more than the held out composition. 
% In contrast, PoE achieves the highest precision on Generalization, and LoRA further improves the balance of coverage and fidelity, reaching $67.8\%$ recall and $58.5\%$ F1 on the Generalization set compared with Query-only at $40.0\%$ and $47.4\%$. 
% This supports the hypothesis that the PoE teacher recovers richer distributional information by composing local neighborhoods found in the score field, while LoRA distillation translates that information into a more sampleable conditional.

% The qualitative results in the CelebA panel of \Cref{fig:dual-panel} ground this interpretation. 
% The PoE and LoRA columns vary identity while preserving the queried attribute combination, whereas the Query-only column repeats near copies of $x_q$. 
% The Top 1 and Top 3 columns drift toward different attribute combinations, showing that composition is not recovered by merging or retrieving the most relevant seen classes. 

% Together, the two datasets support our test-time mechanisms for using discovered modes in a frozen diffusion model, with LoRA adding a pathway for absorbing the discovered composition back into the base score manifold.
On CelebA, the same argument becomes sharper because the images contain greater intra-class diversity and higher pixel-level information density. 
Query-only attains the strongest Faithfulness scores, including $64.0\%$ F1, but its Generalization F1 drops to $20.2\%$. 
This gap indicates that $q_{x_q}$ captures the data point more than the held-out composition. 
In contrast, PoE achieves the best Generalization FID, CLIP, and precision, reaching $140.3$ FID, $68.4\%$ CLIP, and $74.5\%$ precision. 
LoRA further improves the balance of coverage and fidelity, reaching the highest Generalization recall and F1, with $16.8\%$ recall and $24.7\%$ F1 compared with Query-only at $13.4\%$ and $20.2\%$. 
This supports the hypothesis that the PoE teacher recovers richer distributional information by composing local neighborhoods found in the score field, while LoRA distillation translates that information into a more sampleable conditional.
The qualitative results in the CelebA panel of \Cref{fig:dual-panel} ground this interpretation. 
The PoE and LoRA columns vary identity while preserving the queried attribute combination (smiling, blond hair, wavy hair), whereas the Query-only column repeats near copies of $x_q$. 
The Top 1 and Top 3 columns drift toward different attribute combinations, showing that composition is not recovered by merging or retrieving the most relevant seen classes. 

Together, the two datasets support our test-time mechanisms for using discovered modes in a frozen diffusion model, with LoRA adding a pathway for absorbing the discovered composition back into the base score manifold.
\section{Conclusion, Limitation, and Future work}
\label{sec:conclusion}

% We presented a test-time framework that connects concept discovery and compositional generation in pretrained diffusion models. Rather than assuming a fixed primitive library, our method ascends the unconditional DDPM score field to recover query-specific prototypes, maps them into clean-space local Gaussian experts, and composes them with a product-of-experts teacher whose analytic score guides diffusion sampling.
% Empirically, our results support the hypothesis that unseen compositions are partly encoded in the local score geometry around a query. On ColorMNIST and CelebA, PoE composition and its LoRA-distilled variant improve over query-only sampling and nearest trained-class retrieval, producing samples that better balance faithfulness to the query with generalization to other members of the same held-out composition. These findings suggest that mode-found prototypes capture reusable structure beyond point reconstruction or class-level retrieval.
We presented a test-time framework that connects concept discovery and compositional generalization in pretrained diffusion models. Rather than assuming a fixed primitive library, our method uses the diffusion's learned time-indexed scores of the noisy marginals $p_t(x_t)$ to recover query-specific density modes, rescales them into clean-space local Gaussian experts, and composes them with a product-of-experts teacher model whose analytic score guides diffusion sampling. 
Empirically, our results support the hypothesis that unseen compositions are encoded in the local mode geometry of the learned diffusion marginals. On ColorMNIST and CelebA, PoE composition and its LoRA-distilled variant improve over query-only sampling and nearest trained-class retrieval, producing samples that better balance faithfulness to the query with generalization to other members of the same held-out composition. 
These findings suggest that diffusion models encode reusable concept prototypes beyond point reconstruction or class-level retrieval, providing both theoretical and empirical evidence for broader compositional applications, including language modeling~\citep{li2022diffusionlm,nie2026large} and robotic control~\citep{chi2023diffusionpolicy}.

\textbf{Limitation.} A key limitation is that our protocol follows the standard compositional generalization setting: primitive factors are known, while test cases are unseen recombinations. This setting is still meaningful for large foundation models, whose pretraining may already contain a rich space of primitives, making many novel instances closer to unseen compositions than entirely unseen concepts. However, human-level intelligence also requires learning new primitives and incorporating them into future compositions \cite{fodor1988connectionism,lake2017building,lake2015human}. 
Empirically, we observe that our framework can sometimes discover novel-looking primitives by interpolating between known primitives (see Appendix \ref{app:lim}). A full mechanism for detecting, consolidating, and reusing genuinely new primitives remains an important direction for future work.

\bibliography{references}
\bibliographystyle{unsrt}

\appendix
% =============================================================
% Appendix fragment: OOD compositional experiment
% =============================================================
\appendix

\section{Diffusion marginals as a hierarchy of modes}
\label{app:fmnist}
\begin{figure*}[h]
    \centering
    %% Adjust this if your LaTeX source lives elsewhere.
    %% \graphicspath{{concept-decomposition/mode_from_data/}}
    \begin{subfigure}[b]{0.19\textwidth}
        \includegraphics[width=\linewidth]{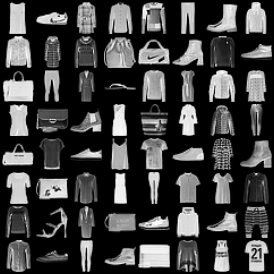}
        \caption{$t=0$}
    \end{subfigure}\hfill
    \begin{subfigure}[b]{0.19\textwidth}
        \includegraphics[width=\linewidth]{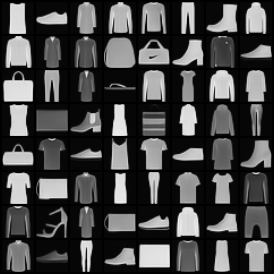}
        \caption{$t=50$}
    \end{subfigure}\hfill
    \begin{subfigure}[b]{0.19\textwidth}
        \includegraphics[width=\linewidth]{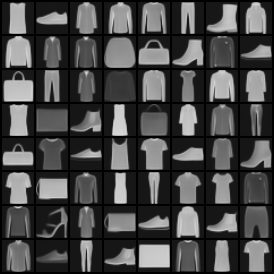}
        \caption{$t=200$}
    \end{subfigure}\hfill
    % \begin{subfigure}[b]{0.085\textwidth}
    %     \includegraphics[width=\linewidth]{figs/fmnist/modes_t150.png}
    %     \caption{$t=150$}
    % \end{subfigure}\hfill
    % \begin{subfigure}[b]{0.085\textwidth}
    %     \includegraphics[width=\linewidth]{figs/fmnist/modes_t150.png}
    %     \caption{$t=200$}
    % \end{subfigure}\hfill
    % \begin{subfigure}[b]{0.085\textwidth}
    %     \includegraphics[width=\linewidth]{figs/fmnist/modes_t200.png}
    %     \caption{$t=250$}
    % \end{subfigure}\hfill
    \begin{subfigure}[b]{0.19\textwidth}
        \includegraphics[width=\linewidth]{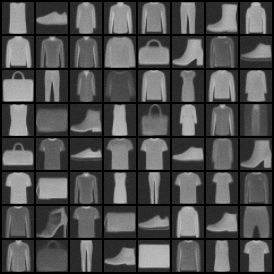}
        \caption{$t=300$}
    \end{subfigure}\hfill
    \begin{subfigure}[b]{0.19\textwidth}
        \includegraphics[width=\linewidth]{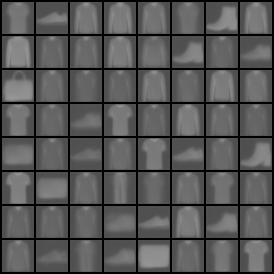}
        \caption{$t=500$}
    \end{subfigure}
    \caption{
    Modes of the noisy marginals $p_t(x_t)$ learned by a DDPM on Fashion-MNIST~\citep{xiao2017fashion}. 
    The $t=0$ panel shows clean reference images, while larger $t$ panels show prototypes recovered by mode ascent at progressively noisier marginals. 
    As $t$ increases, fine instance details are smoothed away and modes consolidate into coarser object-level prototypes, suggesting that diffusion marginals encode an implicit hierarchy of discrete density modes.
    }
\label{fig:fashion-mnist-mode-hierarchy}
    \label{fig:cifar_modes_hierarchical}
\end{figure*}

To motivate our use of mode finding for concept discovery, we visualize local modes of the noisy marginals $p(x_t)$ learned by a DDPM trained on Fashion-MNIST~\citep{xiao2017fashion}. 
Using the mode-ascent procedure in \Cref{sec:hierarchy}, we recover concept prototypes at several noising timesteps. 
As $t$ increases, the marginals become progressively smoother.
Modes that initially preserve instance-level details merge into coarser object groups and eventually retain only high-level class-like structure. 
This supports our view that a diffusion model represents an implicit discrete hierarchy of concepts, where modes of $p(x_t)$ at different noise levels provide prototypes at different levels of abstraction.

\section{Additional benchmark details}
\label{app:benchmark}

\begin{table}[h]
\centering
\caption{Color-MNIST primitive attributes and split.}
\label{tab:cm-primitives}
\small
\begin{tabular}{lc}
\toprule
Attribute & Cardinality \\
\midrule
Digit identity & $10$ \\
Digit color & $4$ \\
Background color & $4$ \\
\midrule
Total compositions & $160$ \\
Seen slots (training) & $120$ \\
Unseen slots (OOD) & $40$ \\
Per-slot images & $1{,}000$ \\
Image format & $32{\times}32$ RGB \\
\bottomrule
\end{tabular}
\end{table}

The Color-MNIST primitive values are digit colors in $\{$yellow, green, cyan, pink$\}$ and background colors in $\{$deep red, navy, dark purple, dark brown$\}$, a muted high-contrast palette chosen so that no digit color collapses against a background under low-light rendering. The $40$ unseen slots are formed by holding out four (digit color, background color) pairs --- (cyan, deep red), (green, navy), (pink, dark purple), (yellow, dark brown) --- across all ten digits, giving $4 \times 10 = 40$ OOD slots; the remaining $120$ slots train the backbone.

\begin{table}[h]
\centering
\caption{CelebA compositional benchmark~\cite{liu2015deep}. A compositional class is a unique combination of $14$ binary CelebA attributes (Black/Blond/Brown/Gray hair, Wavy hair, No\_Beard, Smiling, Big\_Nose, Big\_Lips, Oval\_Face, Male, Sideburns, Wearing\_Necklace, Bald). The $74$ realized combinations used in this experiment are partitioned into a seen tier used for backbone training and a holdout tier accessible only at test time.}
\label{tab:celeba-primitives}
\small
\begin{tabular}{lc}
\toprule
Attribute / split & Value \\
\midrule
Number of binary attributes & $14$ \\
Realized compositional classes & $74$ \\
\midrule
Seen classes (training) & $51$ \\
\quad train images & $26{,}775$ \\
\quad in-distribution validation images & $3{,}825$ \\
Holdout classes (OOD) & $23$ \\
\quad images & $9{,}914$ \\
\midrule
Image format & $128{\times}128$ RGB \\
\bottomrule
\end{tabular}
\end{table}

The $23$ unseen CelebA classes are chosen so that their attribute combination differs from at least one seen class on a single attribute, providing a controlled OOD setting where each held-out class is reachable from a near neighbor in the seen set. Class ids are remapped contiguously to $\{0,\ldots,119\}$ for Color-MNIST and $\{0,\ldots,50\}$ for CelebA before being passed to the backbone's class embedding; the backbone never sees any image from any held-out tier during training. 
% For every unseen class, the deterministic indices $0,\ldots,9$ of the test set are designated the queries $\{x_q^{(i)}\}_{i=1}^{10}$; the remaining images of the same class, with these $10$ indices excluded, form the leak-free real reference bank used to construct the two reference sets of Appendix~\ref{app:eval}. Data loaders and split metadata are released with the code.

\section{Backbone architecture and pretraining}
\label{app:backbone}

The backbone is a class-conditional \texttt{diffusers} \texttt{UNet2DModel}~\cite{ho2020denoising} trained with classifier-free guidance dropout~\cite{ho2022classifier} on top of two off-the-shelf UNet configurations: \texttt{google/ddpm-cifar10-32} for the $32{\times}32$ Color-MNIST backbone, and \texttt{google/ddpm-celebahq-256} (with \texttt{sample\_size} overridden to $128$) for the $128{\times}128$ CelebA backbone. The class-embedding table is sized to $|\mathcal{C}|+1$ to reserve one row for the null token used during CFG dropout. Training uses the DDPM scheduler with a linear $\beta$ schedule, $T = 1000$ training timesteps, $\varepsilon$-prediction, and an EMA on UNet weights. Color-MNIST is trained on a single GPU; CelebA is trained on $4$ GPUs in bf16 mixed precision and we evaluate the EMA checkpoint at training step $500{,}000$. Table~\ref{tab:backbone} lists the per-dataset training hyperparameters; architectural fields not listed (block out-channels, layers-per-block, attention resolutions) follow the base UNet config of each dataset.

\begin{table}[h]
\centering
\caption{Backbone training hyperparameters. Architectural fields not shown follow the base UNet config of each dataset (\texttt{google/ddpm-cifar10-32} or \texttt{google/ddpm-celebahq-256} with \texttt{sample\_size}=$128$).}
\label{tab:backbone}
\small
\begin{tabular}{lcc}
\toprule
Parameter & Color-MNIST & CelebA \\
\midrule
Image resolution & $32$ & $128$ \\
Channels & $3$ & $3$ \\
Base UNet config & \texttt{ddpm-cifar10-32} & \texttt{ddpm-celebahq-256} (\texttt{sample\_size}=$128$) \\
Class-embedding rows (incl.\ null) & $121$ & $52$ \\
Optimizer & AdamW & AdamW \\
Learning rate & $2{\times}10^{-4}$ & $2{\times}10^{-4}$ \\
Weight decay & $10^{-6}$ & $10^{-6}$ \\
Gradient clip ($\|\nabla\|_2$) & $1.0$ & $1.0$ \\
Batch size & $256$ & $128$ ($32$ per GPU $\times 4$) \\
Training length & $100$ epochs & $500{,}000$ steps \\
Mixed precision & bf16 & bf16 \\
EMA on UNet weights & yes & yes \\
Null-token dropout $p$ & $0.1$ & $0.1$ \\
Diffusion schedule & linear $\beta$, $T{=}1000$ & linear $\beta$, $T{=}1000$ \\
Parameterization & $\varepsilon$-prediction & $\varepsilon$-prediction \\
\bottomrule
\end{tabular}
\end{table}

\section{Hyperparameters}
\label{app:hypers}
\subsection{Concept discovery hyperparameters}
\label{app:discovery}
The per-mode covariance $\Sigma_k^{(t_k)}$ is modeled diagonal, and its diagonal is estimated from Hutchinson probes~\cite{hutchinson1990stochastic} of the Tweedie Jacobian. See \Cref{tab:discovery}.
\begin{table}[h!]
\centering
\caption{Concept discovery settings (per dataset). Both datasets use $K{=}3$ prototypes and the per-coordinate softmax weighting of Eq.~\eqref{eq:per-dim-weights} with temperature $\tau$. The differences are mostly in the discovery sampling budget, which is larger on Color-MNIST because per-step ascent on a $32{\times}32$ tensor is much cheaper than on $128{\times}128$.}
\label{tab:discovery}
\small
\begin{tabular}{lcc}
\toprule
Hyperparameter & Color-MNIST & CelebA \\
\midrule
Number of prototypes $K$ & $3$ & $3$ \\
Selection objective & submodular greedy & submodular greedy \\
Per-coordinate softmax temperature $\tau$ & $0.5$ & $0.5$ \\
Timestep grid (\texttt{[start,\,end,\,step]}) & $[50,\,400,\,25]$ & $[50,\,500,\,50]$ \\
Ascent starts per timestep $n_{\text{per-}t}$ & $128$ & $8$ \\
Ascent iterations $n_{\text{iters}}$ & $150$ & $100$ \\
Ascent base step size & $0.1$ & $0.09$ \\
Ascent optimizer & Adam & Adam \\
Ascent initialization & hybrid (50\% query, 50\% random) & query-centered \\
% Selection mode & global-top-$K$ & global-top-$K$ \\
Hutchinson probes & $4$ & $4$ \\
Hutchinson finite-difference step & $5{\times}10^{-3}$ & $5{\times}10^{-3}$ \\
% Per-coordinate weight floor & $0$ & $0$ \\
Unconditional score source & $\emptyset$ token & $\emptyset$ token \\
Seed & $42$ & $42$ \\
\bottomrule
\end{tabular}
\end{table}

% The method section writes the score-ascent objective as plain gradient ascent; we use Adam in practice for numerical stability, and the two optimizers produce qualitatively identical modes in our setting. 

\subsection{Distillation hyperparameters}
\label{app:distill}

\begin{table}[h]
\centering
\caption{Distillation settings (per dataset). The trainable parameters are jointly the LoRA adapter on the frozen UNet and the new class embedding $c_{\text{new}}$, optimized on a fixed pool $\mathcal{D}_T$ of $N_{\text{pool}}$ images sampled from the analytic teacher under the variance-aware schedule (Appendix~\ref{app:sampling}).}
\label{tab:distill}
\small
\begin{tabular}{lcc}
\toprule
Hyperparameter & Color-MNIST & CelebA \\
\midrule
\multicolumn{3}{l}{\textit{LoRA adapter on the frozen UNet}} \\
LoRA rank $r$ & $8$ & $16$ \\
LoRA target modules & full UNet linear/conv & full UNet linear/conv \\
LoRA $\alpha$ & $2r{=}16$ & $2r{=}32$ \\
LoRA dropout & $0.0$ & $0.0$ \\
LoRA learning rate & $1{\times}10^{-3}$ & $1{\times}10^{-3}$ \\
\midrule
\multicolumn{3}{l}{\textit{New class embedding $c_{\text{new}}$}} \\
Embedding dim $d_c$ & matches backbone class-embed dim & matches backbone class-embed dim \\
Initialization & $\mathcal{N}(0,\,0.01^2)$ & $\mathcal{N}(0,\,0.01^2)$ \\
Learning rate & $1{\times}10^{-3}$ & $1{\times}10^{-3}$ \\
\midrule
\multicolumn{3}{l}{\textit{Optimization}} \\
Optimizer & Adam & Adam \\
Mini-batch size & $16$ & $2$ \\
Epochs & $10$ & $5$ \\
CFG dropout on $c_{\text{new}}$ & $p{=}0.1$ & $p{=}0.1$ \\
\midrule
\multicolumn{3}{l}{\textit{Pool $\mathcal{D}_T$ used as distillation target}} \\
Pool size $N_{\text{pool}}$ & $256$ & $256$ \\
Pool sampler & DDIM, $\eta{=}0$, $50$ steps & DDIM, $\eta{=}0$, $50$ steps \\
Pool guidance schedule & variance-aware (Eq.~\ref{eq:variance-guidance}) & variance-aware (Eq.~\ref{eq:variance-guidance}) \\
\bottomrule
\end{tabular}
\end{table}

The trainable parameters are the LoRA factors $\theta_{\text{LoRA}}$ inserted on every linear and convolutional layer of the UNet plus the single class embedding $c_{\text{new}}$. The base UNet weights and the backbone's class-embedding table are frozen throughout. The training data is the fixed pool $\mathcal{D}_T = \{x_0^{(i)}\}_{i=1}^{N_{\text{pool}}}$ drawn once via the analytic compositional sampler of Eq.~\ref{eq:comp-plusqT} under the variance-aware schedule of Eq.~\ref{eq:variance-guidance}; each Adam step samples a mini-batch from $\mathcal{D}_T$, applies a random forward-noising at $t \sim \mathcal{U}\{1,\ldots,T\}$, and minimizes the standard $\varepsilon$-prediction loss of Eq.~\ref{eq:lora-distill-loss} jointly on $(\theta_{\text{LoRA}},\,c_{\text{new}})$. CFG dropout on $c_{\text{new}}$ at $p{=}0.1$ keeps the unconditional branch $\varepsilon_{\theta+\theta'}(x_t,t,\emptyset)$ usable at inference. Sampling along the model's score field (rather than from raw analytic Gaussian draws of $q_T$) keeps the distillation target on the learned image manifold. A separate $(\theta_{\text{LoRA}},\,c_{\text{new}})$ pair is distilled per OOD query.

\subsection{Sampling hyperparameters}
\label{app:sampling}

\begin{table}[h]
\centering
\caption{Sampling settings used for both pool generation (training $\mathcal{D}_T$ for distillation, and the analytic-PoE evaluation pool) and inference from the distilled model. The variance-aware $w_0,\,w_{\max}$ control the strength of analytic-PoE guidance during pool generation. The LoRA sampling-$w$ controls standard CFG strength when sampling from $c_{\text{new}}$ after distillation; we report a sweep around the canonical value.}
\label{tab:sampling}
\small
\begin{tabular}{lcc}
\toprule
Hyperparameter & Color-MNIST & CelebA \\
\midrule
DDIM scheduler & $\eta{=}0$, $50$ steps & $\eta{=}0$, $50$ steps \\
Pool size (analytic-PoE sampler) & $256$ & $256$ \\
Variance-aware base scale $w_0$ & $1.2$ & $0.5$ \\
Variance-aware cap $w_{\max}$ & $2.0$ & $2.0$ \\
Variance-mode & entropy-adaptive & entropy-adaptive \\
LoRA sampling-$w$ (canonical) & $0.8$ & $1.2$ \\
LoRA sampling-$w$ sweep & $\{0.8,\,1.0,\,1.2\}$ & $\{0.8,\,1.0,\,1.2\}$ \\
Saved gens per query per teacher & $16$ & $16$ \\
Seed (fixed across methods) & $42$ & $42$ \\
\bottomrule
\end{tabular}
\end{table}

\section{Additional evaluation details}
\label{app:eval}

\paragraph{Small-$N$ caveats.}
At $N{=}100$, FID's covariance term is poorly estimated and the metric is dominated by the squared-mean distance. The estimator is therefore biased upward but the bias is shared across methods at the same $N$, so within-table comparisons remain valid; absolute FID values should not be compared against published $N{=}50,000$ FIDs. 
% Similarly, the $k$-NN radii used by precision/recall are computed from the within-set distances and are noisy at small $N$. We report mean $\pm$ standard error across the OOD-class population, and recommend reading any single-class number with the per-class scatter.

\section{Compute}
\label{app:compute}
Backbone training uses a single NVIDIA A40 GPU for ColorMNIST,
completing in roughly 6 hours, and four NVIDIA RTX 6000 GPUs for
CelebA, completing in roughly 50 hours. Test time concept discovery
and PoE sampling both run on a single A40, and the full pipeline of
\cref{sec:method} (mode ascent, Hutchinson estimation of the local
Hessian, submodular selection, PoE composition, and analytic CFG
sampling) takes about 10-17 minutes per query.

\section{Visualizations}
\label{app:viz}
\begin{figure*}[h]
    \centering

    \begin{minipage}{1\textwidth}
        \centering
        \includegraphics[width=\linewidth]{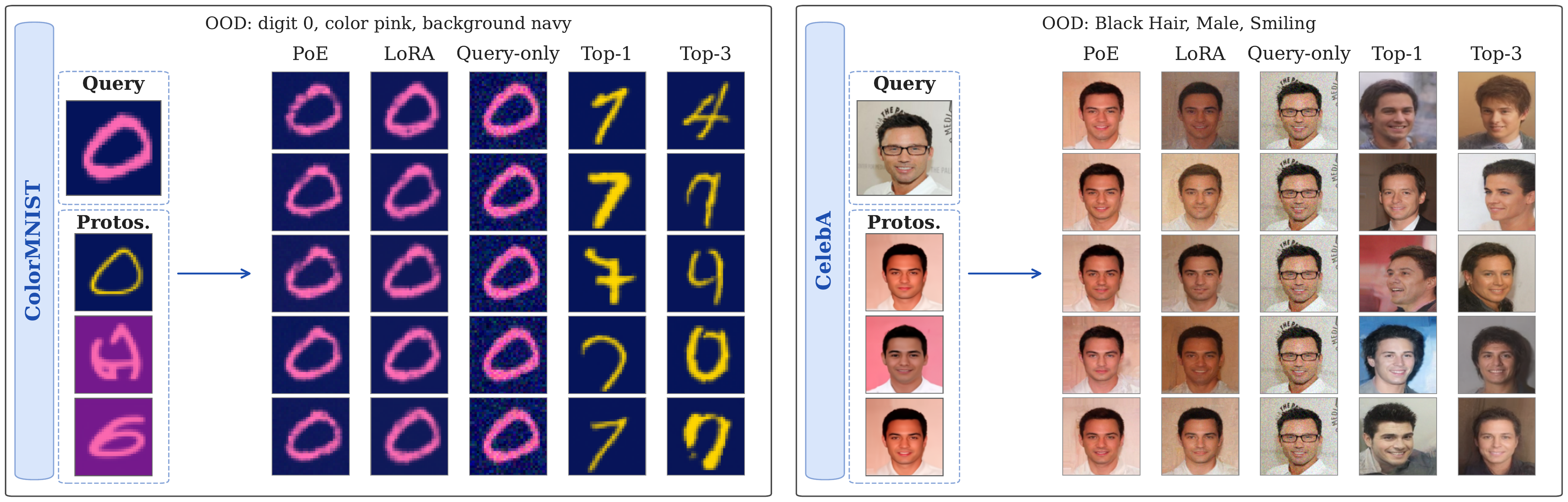}
        \vspace{-0.5em}
        % \caption*{(a) Additional qualitative results, part 1.}
    \end{minipage}

    \vspace{1em}

    \begin{minipage}{1\textwidth}
        \centering
        \includegraphics[width=\linewidth]{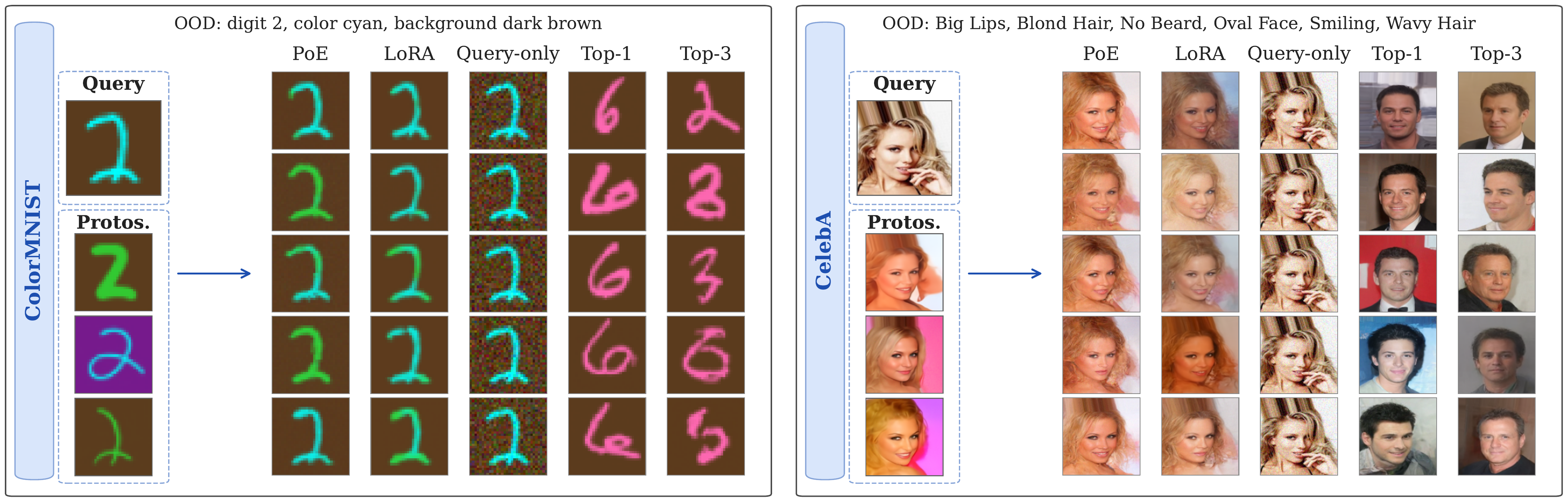}
        \vspace{-0.5em}
        % \caption*{(b) Additional qualitative results, part 2.}
    \end{minipage}

    \vspace{1em}

    \begin{minipage}{1\textwidth}
        \centering
        \includegraphics[width=\linewidth]{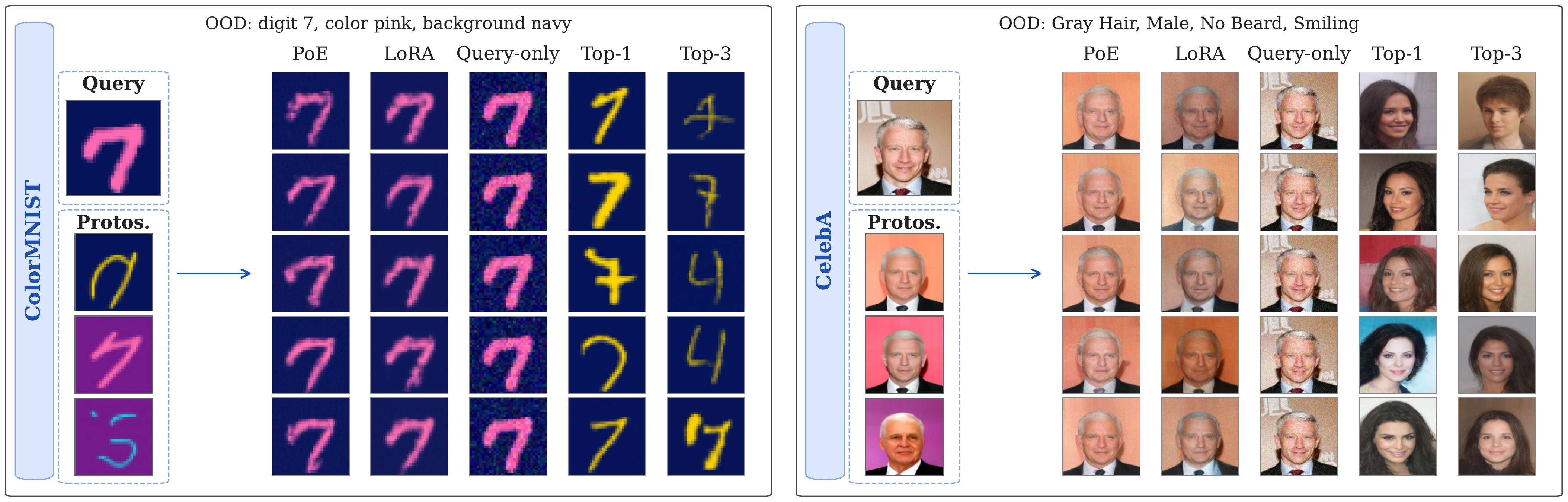}
        \vspace{-0.5em}
        % \caption*{(c) Additional qualitative results, part 3.}
    \end{minipage}

    \caption{Additional qualitative results for the ColorMNIST and CelebA.}
    \label{fig:dual-dataset-appendix}
\end{figure*}

We present additional visualization for both datasets in \Cref{fig:dual-dataset-appendix}.
% TODO: qualitative failure cases and prototype visualizations

\section{Generalization to unseen primitives.}
\begin{figure}
    \centering
    \includegraphics[width=0.4\linewidth]{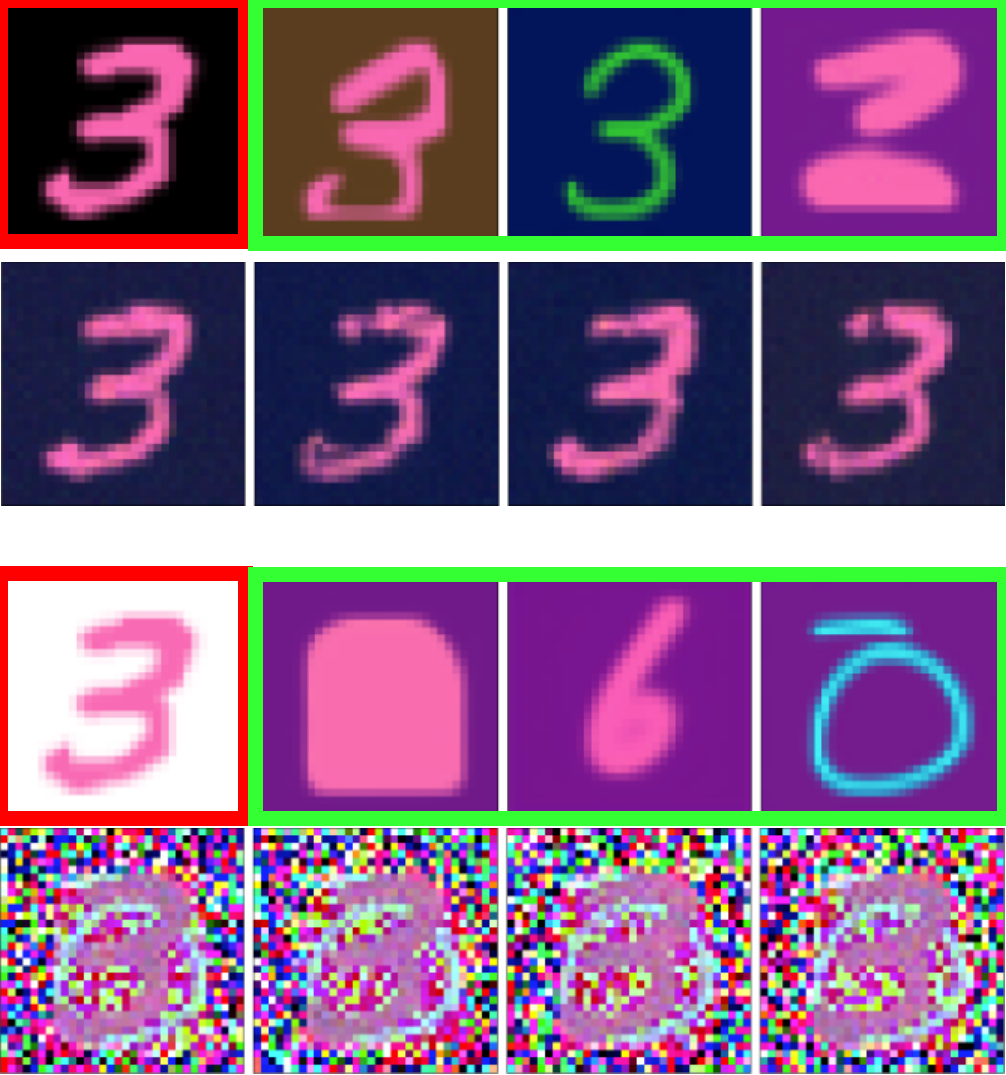}
    \caption{
Concept discovery on novel background primitives, pink digit fixed from ColorMNIST.
Each panel shows the query (red border), three discovered prototypes
(green borders, top row), and four PoE samples (bottom row).
\textbf{Top:} pink digit on a black background, an interpolation
inside the convex hull of the trained backgrounds. The discovered
prototypes and PoE samples reproduce the held out background as a
coherent novel primitive. \textbf{Bottom:} the same digit on a white
background, an extrapolation far outside the trained hull. The
prototypes drift to trained dark backgrounds and the samples collapse
to salt and pepper texture, evidence that the score field carries no
information about the held out brightness.
}
    \label{fig:novel_primitives_pink3}
\end{figure}
\label{app:lim}
Concept discovery operates on the score field induced by the trained DDPM,
so the prototypes it can surface are confined to regions where that score
field is informative. \Cref{fig:novel_primitives_pink3} illustrates this
boundary with two queries that share an in palette pink digit but differ in
background. The first places the digit on a black background, which lies
inside the convex hull of the four trained backgrounds (\textit{deep red},
\textit{navy}, \textit{dark purple}, \textit{dark brown}), all of which are
dark and unsaturated. The second places the digit on a white background,
which falls far outside that hull. The discovered prototypes and the PoE
samples reproduce the held out black background as a coherent novel
primitive, while the white query collapses onto trained dark backgrounds
with the pink digit preserved but the brightness of the background lost.
The discovery procedure therefore inherits the inductive biases of the
backbone: it can compose new primitives that interpolate between seen ones,
but cannot fabricate primitives that extrapolate beyond the training
support.

\section{Broader Impacts}
\label{app:broader}
A potential positive impact is reducing the data and compute needed to 
generate novel visual concepts at test time. A potential negative impact 
is that the same capability could lower the barrier to generating face 
images with novel attribute combinations, which could be misused for 
misleading imagery. However, the method operates on small-scale 
class-conditional DDPMs and does not approach the fidelity of modern 
text-to-image systems.

\section{Licenses for Existing Assets}
\label{app:licenses}

\begin{table}[h]
\centering
\caption{Licenses for existing assets.}
\label{tab:licenses}
\begin{tabular}{lll}
\toprule
Asset & License & Source \\
\midrule
MNIST & CC BY-SA 3.0 & \href{http://yann.lecun.com/exdb/mnist/}{MNIST website} \\
CelebA & Non-commercial research & \href{https://mmlab.ie.cuhk.edu.hk/projects/CelebA.html}{CelebA website} \\
HF diffusers & Apache 2.0 & \href{https://github.com/huggingface/diffusers}{GitHub} \\
\bottomrule
\end{tabular}
\end{table}

% \newpage
% \input{checklist.tex}

\end{document}